\documentclass[letterpaper, 10 pt, conference]{ieeeconf}  

\usepackage{amsmath,amsfonts}
\usepackage{algorithmic}
\usepackage{array}
\usepackage{amssymb} 
\usepackage{adjustbox} 
\usepackage{textcomp}
\usepackage{stfloats}
\usepackage{subcaption}
\usepackage{pifont}
\usepackage{cite}
\usepackage{makecell} 

\makeatletter
\let\NAT@parse\undefined
\makeatother
\usepackage{lineno,hyperref}
\usepackage{booktabs} 
\usepackage{svg}
\usepackage{graphicx}
\usepackage{subcaption}
\usepackage{multirow}
\usepackage{booktabs}
\usepackage{makecell} 
\hypersetup{
	colorlinks=true, 
	linkcolor=blue,  
	citecolor=blue, 
	urlcolor=red     
}
\usepackage{url}
\usepackage{verbatim}
\usepackage{graphicx}
\usepackage{xcolor}
\def\BibTeX{{\rm B\kern-.05em{\sc i\kern-.025em b}\kern-.08em
		T\kern-.1667em\lower.7ex\hbox{E}\kern-.125emX}}
\usepackage{balance}

\title{\LARGE \bf
A Flow Matching Framework for Soft‑Robot Inverse Dynamics}

\author{Hang Yang$^{\dagger}$, Fangju Yang$^{\dagger}$, Yangming Zhang$^{}$, Ibrahim Alsarraj$^{}$, Yuhao Wang$^{}$, Zhenye Luo$^{}$, Zixi Chen$^{}$, Ke Wu$^{*}$ 
\thanks{}
}

\begin{document}

\maketitle
\thispagestyle{empty}
\pagestyle{empty}

\begin{abstract}
Learning the inverse dynamics of soft continuum robots remains challenging due to high-dimensional nonlinearities and complex actuation coupling. Conventional feedback-based controllers often suffer from control chattering due to corrective oscillations, while deterministic regression-based learners struggle to capture the complex nonlinear mappings required for accurate dynamic tracking. Motivated by these limitations, we propose an inverse-dynamics framework for open-loop feedforward control that learns the system’s differential dynamics as a generative transport map. Specifically, inverse dynamics is reformulated as a conditional flow-matching problem, and Rectified Flow (RF) is adopted as a lightweight instance to generate physically consistent control inputs rather than conditional averages. Two variants are introduced to further enhance physical consistency: RF-Physical, utilizing a physics-based prior for residual modeling; and RF-FWD, integrating a forward-dynamics consistency loss during flow matching. Extensive evaluations demonstrate that our framework reduces trajectory tracking RMSE by over 50\% compared to standard regression baselines (MLP, LSTM, Transformer). The system sustains stable open-loop execution at a peak end-effector velocity of 1.14 m/s with sub-millisecond inference latency (0.995 ms). This work demonstrates flow matching as a robust, high-performance paradigm for learning differential inverse dynamics in soft robotic systems.
\end{abstract}


\section{Introduction}
\label{sec:introduction}

Soft continuum robots are attractive for safe interaction and adaptive motion because compliance enables large deformation and passive accommodation of contact \cite{rus2015design,thuruthel2018survey}. The same compliance, however, introduces hysteresis, inertia, and strong actuation--state coupling \cite{thuruthel2020first,liu2025review}, so accurate dynamic tracking becomes difficult once inertial effects become pronounced \cite{thuruthel2018survey,liu2025review}. Existing control strategies are therefore usually discussed under two broad categories, namely model-based and model-free controllers \cite{thuruthel2018survey,liu2025review}. Across both categories, actuation is still most often generated within feedback-controller architectures, where sensed state or tracking error is used to update the command online \cite{haggerty2023inertial,licher2025adaptive}. When such feedback is unavailable, delayed, or undesirable, there remains a need for a lightweight inverse-dynamics mapping for direct open-loop feedforward execution\cite{thuruthel2018openloop,thuruthel2020first}.

Model-based control methods derive actuation from explicit descriptions of soft-body mechanics \cite{thuruthel2018survey,liu2025review}. Existing research in this field is predominantly driven by two control paradigms. The first is rooted in directly designing controllers from physics-based models, which encompasses continuum mechanics formulations (e.g., finite element methods \cite{wu2022fem}), Cosserat rod models \cite{mathew2025reduced}, and simplified geometric representations such as piecewise constant curvature models \cite{5PHDchirikjian1992theory}. These models are widely used to construct feedback or inverse-dynamics control laws \cite{thuruthel2018survey}. The second class consists of model-based optimization approaches, such as model predictive control (MPC), which embed system dynamics into online optimization to improve dynamic performance \cite{haggerty2023inertial,licher2025adaptive}. However, higher model fidelity increases reliance on parameter identification, state estimation, and online computation, limiting applicability in dynamic soft-robot control \cite{doroudchi2021inverse,licher2025adaptive}.


Model-free controllers instead learn control-relevant mappings directly from data \cite{liu2025review,chen2024review}. The simplest line uses Jacobian-based relations for inverse kinematics and feedback correction \cite{fang2020ik}. Such methods are concise and easy to implement, but their local linearization becomes inaccurate under strong nonlinearity and hysteresis, which in turn necessitates frequent online updating \cite{liu2025review,fang2020ik}. Neural models then replace local Jacobians with nonlinear input--output mappings: recurrent models learn temporal dynamics \cite{tariverdi2021rnn}, differentiable models support trajectory optimization \cite{bern2020differentiable}, and Koopman-based learning methods provide control-oriented latent dynamics \cite{bruder2021koopmancontrol,feizi2025deepkoopman}. This progression improves expressive power, but the learned model still usually serves prediction, latent lifting, or feedback control rather than direct inverse-dynamics generation \cite{bern2020differentiable,bruder2021koopmancontrol}. A smaller line of work learns the inverse relation more directly, including inverse statics \cite{giorelli2015inversestatics}, actuator-specific inverse dynamics \cite{baysal2022pam,zhang2023gru}, and task-specific inverse control \cite{bianchi2023throwing}. Yet these formulations remain local, static, mechanism-specific, or task-specific, and thus do not provide a general feedforward differential inverse-dynamics map for open-loop trajectory tracking \cite{giorelli2015inversestatics,bianchi2023throwing}. This limitation makes generative modeling attractive, since it can represent distributions over feasible control inputs rather than collapsing them into a single deterministic estimate. Recent generative action models such as diffusion policies can represent complex control distributions \cite{chi2023diffusionpolicy}, but their iterative denoising process remains computationally heavy, which makes real-time soft-robot deployment difficult.

Motivated by these limitations, this work develops a lightweight inverse-dynamics framework for open-loop feedforward control; it learns the actuation associated with local differential state transitions instead of computing commands from instantaneous tracking errors\cite{thuruthel2018openloop,thuruthel2020first}. Specifically, inverse dynamics is reformulated as a conditional flow-matching problem and instantiated with Rectified Flow (RF) \cite{liu2023rectifiedflow,lipman2023flowmatching}. Building on this formulation, the framework is further extended to Rectified Flow with a physical prior (RF-Physical) and Rectified Flow with forward-dynamics consistency (RF-FWD), providing a lightweight feedforward solution for open-loop trajectory tracking in soft continuum robots. The main contributions of this work are:
\begin{itemize}
    \item \textbf{Framework.} A lightweight flow-matching framework for soft-robot inverse dynamics, enabling open-loop feedforward control by mapping local state transitions to actuation inputs via Rectified Flow.
    \item \textbf{Method.} Two physics-consistent variants, RF-Physical and RF-FWD, incorporating residual physical priors and forward-dynamics consistency to improve open-loop rollout stability.
    \item \textbf{Evidence.} Simulation and experiments demonstrate more than 50\% RMSE reduction relative to MLP-, LSTM-, and Transformer-based baselines, with 5.0\,mm RMSE on structured trajectories (1.1\% of the robot length), 0.995\,ms inference time, and stable tracking up to 1.14\,m/s in experiments.
\end{itemize}

\begin{figure*}[t]
    \centering
    \includegraphics[width=0.8\linewidth]{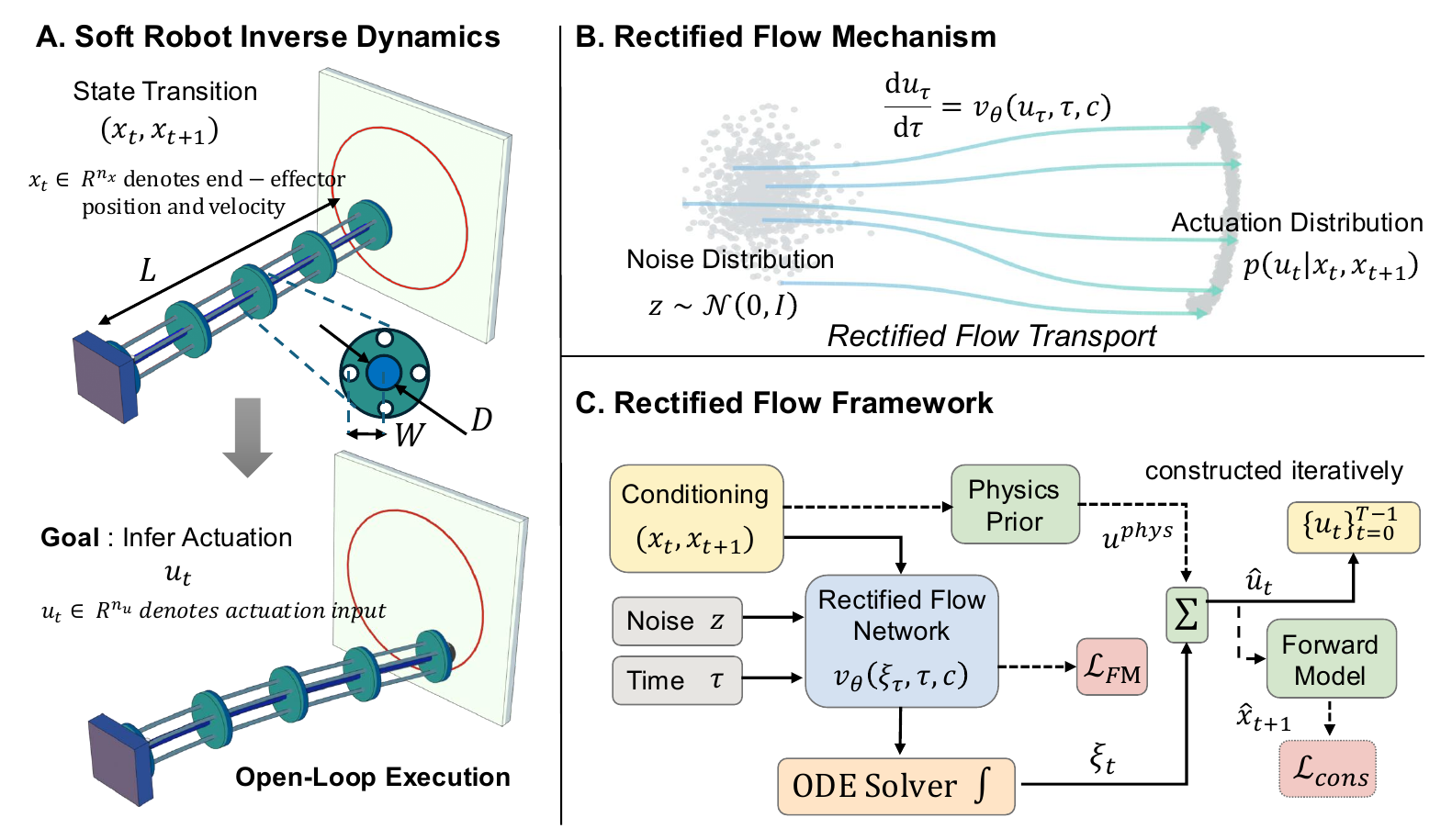}
    \caption{{Flow-matching-based open-loop feedforward controller. The framework solves the inverse-dynamics problem by iteratively mapping desired state transitions to control inputs through learned flow matching.}}
    \label{fig:method_overview}
\end{figure*}
     
\section{PROBLEM STATEMENT}

\subsection{Task Objective and Robot Platform}
We consider open-loop trajectory tracking of a single-segment cable-driven continuum robot in a three-dimensional task space. The robot consists of an elastic beam, 3D-printed spacer disks, and uniformly distributed cables threaded through the disks. With one end fixed, cable actuation produces continuous bending of the compliant body, which can be described by continuum-robot formulations such as Cosserat rod theory \cite{cosserat1909theorie,till2019real}. Given the current state $x_t$ and a desired next state $x_{t+1}$, the objective is to determine the actuation input $u_t$ that realizes this local transition over one control interval, as illustrated in Fig.~\ref{fig:method_overview}A. This is an inverse-dynamics problem for open-loop control.

\subsection{Main Challenges}
This problem remains challenging because the actuation-to-state mapping is highly nonlinear and history-dependent under large deformation,  and inertial effects \cite{trivedi2008geom,thuruthel2020first}. Moreover, under open-loop execution, inverse errors cannot be corrected online and may accumulate during rollout \cite{thuruthel2018openloop,thuruthel2020first}.

\section{METHODOLOGY}

This section presents the proposed Rectified-Flow-based framework for learning open-loop inverse dynamics of soft continuum robots. Starting from a differential description of the robot dynamics, the control problem is reformulated as iterative   inversion between adjacent states. The resulting discrete inverse map is then learned with flow matching and further regularized by forward-dynamics consistency and a residual physics prior.

\subsection{Problem Formulation: From Differential Dynamics to Iterative Inverse Mapping}
Consider the state $x(t)\in\mathbb{R}^{n_x}$ and actuation $u(t)\in\mathbb{R}^{n_u}$ of the studied cable-driven continuum robot. Its dynamics is written in first-order form as
\begin{equation}
    \dot{x}(t)=f(x(t),u(t)),
\end{equation}
where $f(\cdot)$ denotes the nonlinear differential dynamics. For soft continuum robots, this mapping is strongly nonlinear and history-dependent because distributed compliance, damping, actuation coupling, and inertial effects jointly determine the transient response \cite{haggerty2023inertial}.

Over one sampling interval $\Delta t$, the continuous dynamics induce a discrete flow map
\begin{equation}
    x_{t+1}=F_{\Delta t}(x_t,u_t),
\end{equation}
where $F_{\Delta t}(\cdot)$ denotes the state transition generated by $f(\cdot)$. In principle, inverse dynamics seeks the actuation that realizes a desired motion. However, a convenient closed-form inverse with respect to $u_t$ is generally unavailable for the studied soft robot. Instead of analytically inverting the continuous-time dynamics, we directly learn the   discrete inverse map
\begin{equation}
    u_t=g(x_t,x_{t+1}),
    \label{diff_def}
\end{equation}
which returns the actuation required to realize the local transition from $x_t$ to $x_{t+1}$. In this sense, the proposed model learns the inverse of the differential dynamics over one control interval, i.e., differential inverse dynamics.

Given a desired reference trajectory $\{x_t^\star\}_{t=0}^{T}$, the open-loop control sequence is constructed iteratively as
\begin{equation}
    u_t=g(x_t^\star,x_{t+1}^\star), \quad t=0,\ldots,T-1,
\end{equation}
and the resulting $\{u_t\}_{t=0}^{T-1}$ is executed in open loop. Therefore, the trajectory-level control profile is assembled by repeated local inversion along successive state pairs rather than by repeated feedback correction \cite{thuruthel2018survey}.

\subsection{Rectified-Flow-Based Inverse-Dynamics Framework}
As shown in Fig.~\ref{fig:method_overview}B--C, we formulate inverse dynamics as a conditional generative transport problem. Given the transition context $c_t=(x_t,x_{t+1})$, Rectified Flow learns a conditional velocity field that transports a Gaussian sample $z\sim\mathcal{N}(0,I)$ toward the target control over the normalized flow time $\tau\in[0,1]$ \cite{liu2023rectifiedflow,lipman2023flowmatching}. For RF and RF-FWD, the transported variable is the actuation itself, governed by
\begin{equation}
    \frac{\mathrm{d}u_{\tau}}{\mathrm{d}\tau}
    =
    v_{\theta}(u_{\tau},\tau,x_t,x_{t+1}),
    \qquad \tau\in[0,1],
\end{equation}
with the straight-line interpolation
\begin{equation}
    u_{\tau}=z+\tau(u_t-z).
\end{equation}
For RF-Physical, a quasi-static prior $u_t^{phys}$ is first computed from a geometrically exact variable-strain static model of the studied cable-driven continuum robot \cite{renda_gvs_static_2020,mathew2025reduced}:
\begin{equation}
    u_t^{phys}=h_{stat}(x_{t+1}^{cfg}),
\end{equation}
where $x_{t+1}^{cfg}$ denotes the configuration-related component of the desired next state. The model then learns only the residual actuation
\begin{equation}
    \eta_t=u_t-u_t^{phys},
\end{equation}
with transport path
\begin{equation}
    \eta_{\tau}=z+\tau(\eta_t-z).
\end{equation}
The conditional velocity field $v_\theta$ is parameterized by a lightweight multilayer perceptron. For RF and RF-FWD, the input is $[u_\tau,\tau,x_t,x_{t+1}]$, whereas for RF-Physical it becomes $[\eta_\tau,\tau,x_t,x_{t+1},u_t^{phys}]$. In RF-FWD, the forward-dynamics surrogate $\hat{f}_\phi$ is implemented as a lightweight two-layer MLP with 128 hidden units and ReLU activations.

\subsubsection{Training Loss}
We formulate all RF-based variants under a unified conditional flow-matching framework. The overall objective is
\begin{equation}
    \mathcal{L}
    =
    \mathcal{L}_{FM}
    +
    \lambda_{cons}\mathcal{L}_{cons},
\end{equation}
where $\mathcal{L}_{FM}$ denotes the main flow-matching term and $\mathcal{L}_{cons}$ is an optional forward-dynamics consistency regularizer.

\textit{Main flow-matching term:}
Let $\xi_t$ denote the target and $c_t$ the conditioning variable. We define
\begin{equation}
\begin{aligned}
    \mathcal{L}_{FM}
    &=
    \mathbb{E}_{\xi_t,z,\tau}
    \left[
    \left\|
    v_{\theta}(\xi_{\tau},\tau,c_t)-(\xi_t-z)
    \right\|^2
    \right], \\
    \text{with} \quad \xi_{\tau}
    &=
    (1-\tau)z+\tau\xi_t .
\end{aligned}
\end{equation}
For RF and RF-FWD,
\begin{equation}
    \xi_t=u_t,
    \qquad
    c_t=[x_t,x_{t+1}],
\end{equation}
\noindent yielding
\begin{equation}
    \mathcal{L}_{RF}
    =
    \mathbb{E}_{u_t,z,\tau}
    \left[
    \left\|
    v_{\theta}(u_{\tau},\tau,x_t,x_{t+1})-(u_t-z)
    \right\|^2
    \right].
\end{equation}

For RF-Physical,
\begin{equation}
    \xi_t=\eta_t=u_t-u_t^{phys},
    \qquad
    c_t=[x_t,x_{t+1},u_t^{phys}],
\end{equation}
\noindent yielding
\begin{equation}
\begin{aligned}
    \mathcal{L}_{phys}
    =
    \mathbb{E}_{\eta_t,z,\tau}\big[
    &\|v_{\theta}(\eta_{\tau},\tau,x_t,x_{t+1},u_t^{phys}) \\
    &-(\eta_t-z)\|^2
    \big].
\end{aligned}
\end{equation}

\textit{Forward-dynamics consistency loss:}
To regularize the generated inverse predictions, we introduce a learned forward-dynamics surrogate
\begin{equation}
    \hat{x}_{t+1}=\hat{f}_{\phi}(x_t,\hat{u}_t),
\end{equation}
where $\hat{u}_t$ denotes the control generated by the inverse RF model. The consistency loss is defined as
\begin{equation}
    \mathcal{L}_{cons}
    =
    \mathbb{E}_{x_t,x_{t+1}}
    \left[
    \left\|
    \hat{f}_{\phi}(x_t,\hat{u}_t)-x_{t+1}
    \right\|^2
    \right].
\end{equation}

As a result, RF and RF-FWD learn the direct actuation distribution, while RF-Physical learns only the residual around a physically plausible quasi-static estimate.

\subsubsection{Feedforward Control Inference}
At deployment, the inverse model is queried for each local state transition. For RF and RF-FWD, given a Gaussian sample $z\sim\mathcal{N}(0,I)$, the actuation is obtained by integrating the learned flow:
\begin{equation}
    \hat{u}_t
    =
    z+\int_{0}^{1}v_{\theta}(u_{\tau},\tau,x_t,x_{t+1})\,\mathrm{d}\tau.
\end{equation}
For RF-Physical, the model first generates the residual actuation
\begin{equation}
    \hat{\eta}_t
    =
    z+\int_{0}^{1}v_{\theta}(\eta_{\tau},\tau,x_t,x_{t+1},u_t^{phys})\,\mathrm{d}\tau,
\end{equation}
and the final control input is recovered as
\begin{equation}
    \hat{u}_t=u_t^{phys}+\hat{\eta}_t.
\end{equation}
In practice, the transport ODE is numerically integrated with Euler steps. Given a reference trajectory $\{x_t^\star\}_{t=0}^{T}$, the full open-loop actuation sequence is generated sequentially as $\hat{u}_t=g(x_t^\star,x_{t+1}^\star)$ for $t=0,\ldots,T-1$.

\begin{figure*}[]
    \centering
    \begin{subfigure}{0.245\textwidth}
        \includegraphics[width=\linewidth]{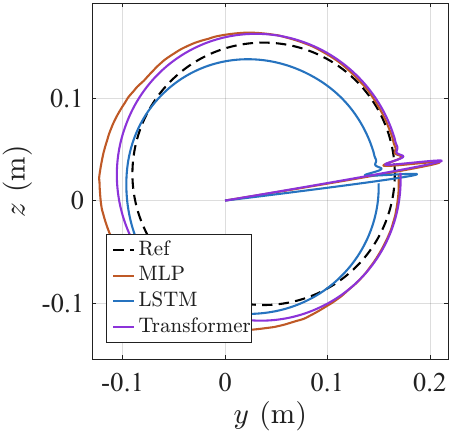}
        \caption{\footnotesize{Circle (MLP/LSTM/Trans)}}
    \end{subfigure}
    \begin{subfigure}{0.245\textwidth}
\includegraphics[width=\linewidth]{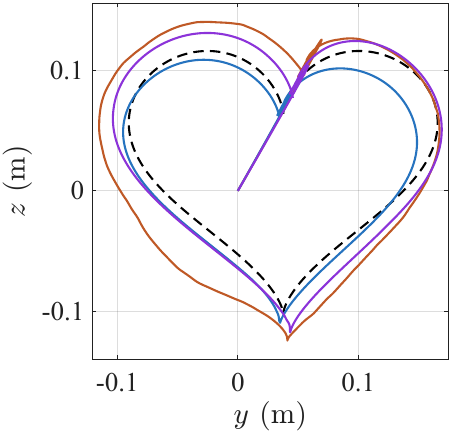}
        \caption{\footnotesize{Heart}}
    \end{subfigure}
    \begin{subfigure}{0.24\textwidth}
        \includegraphics[width=\linewidth]{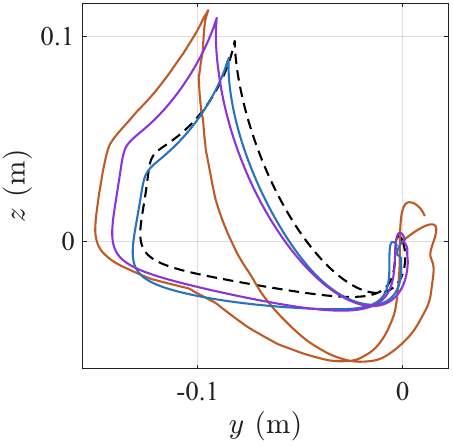}
        \caption{\footnotesize{Rand2}}
    \end{subfigure}
    \begin{subfigure}{0.245\textwidth}
        \includegraphics[width=\linewidth]{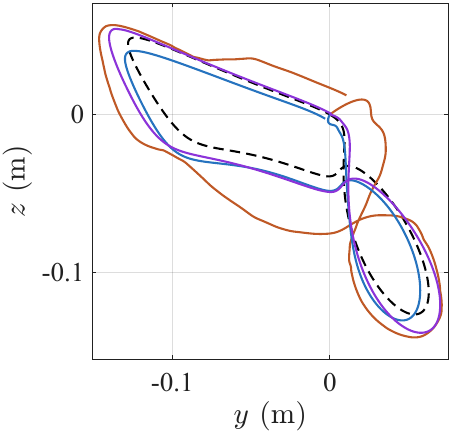}
        \caption{\footnotesize{Rand3}}
    \end{subfigure}
    \begin{subfigure}{0.245\textwidth}
        \includegraphics[width=\linewidth]{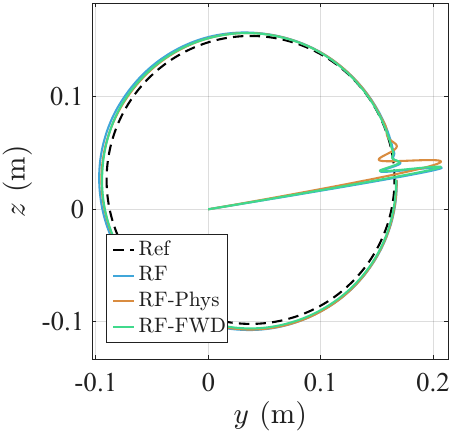}
        \caption{\footnotesize{Circle (RF/RF-Phys/RF-FWD)}}
    \end{subfigure}
    \begin{subfigure}{0.245\textwidth}
        \includegraphics[width=\linewidth]{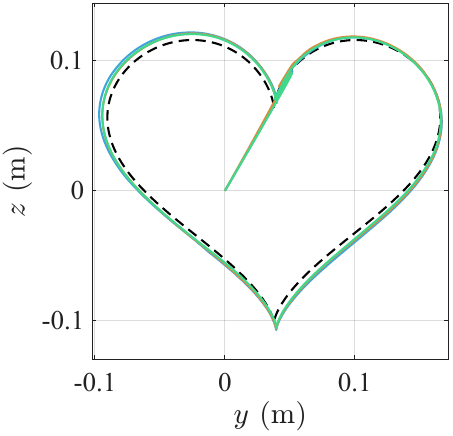}
        \caption{\footnotesize{Heart}}
    \end{subfigure}
    \begin{subfigure}{0.24\textwidth}
        \includegraphics[width=\linewidth]{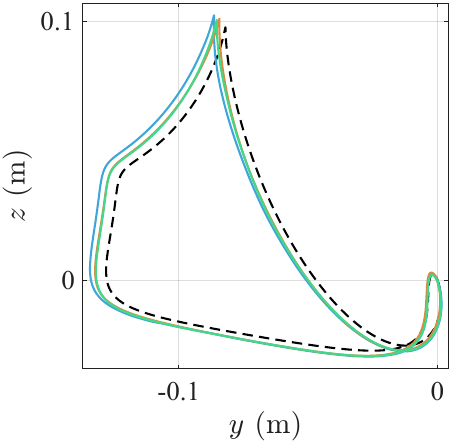}
        \caption{\footnotesize{Rand2}}
    \end{subfigure}
    \begin{subfigure}{0.245\textwidth}
        \includegraphics[width=\linewidth]{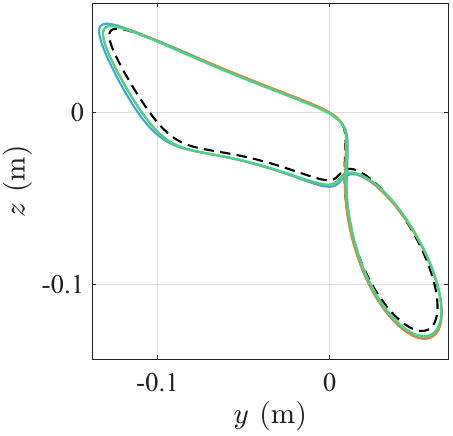}
        \caption{\footnotesize{Rand3}}
    \end{subfigure}
    \caption{\small{Trajectory ($y-z$) comparison. Top: Inverse MLP/LSTM/Transformer. Bottom: RF/RF-Phys/RF-FWD.}}
    \label{fig:tracking_examples}
\end{figure*}

\begin{figure*}[]
    \centering
    \begin{subfigure}{0.245\textwidth}
        \includegraphics[width=\linewidth]{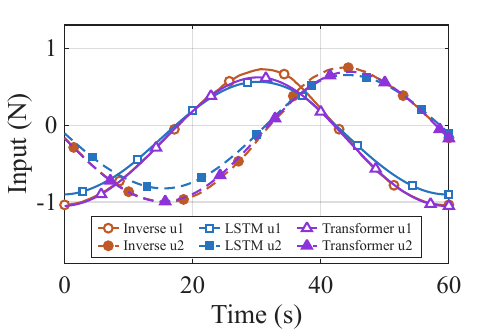}
        \caption{\footnotesize{Circle (MLP/LSTM/Trans)}}
    \end{subfigure}
    \begin{subfigure}{0.245\textwidth}
        \includegraphics[width=\linewidth]{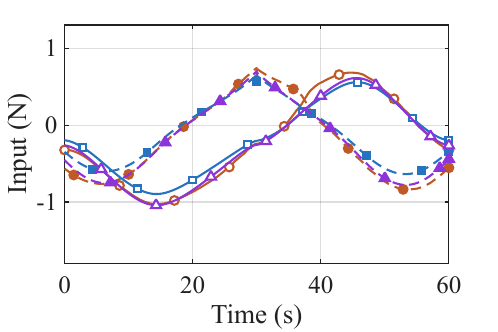}
        \caption{\footnotesize{Heart}}
    \end{subfigure}
    \begin{subfigure}{0.245\textwidth}
        \includegraphics[width=\linewidth]{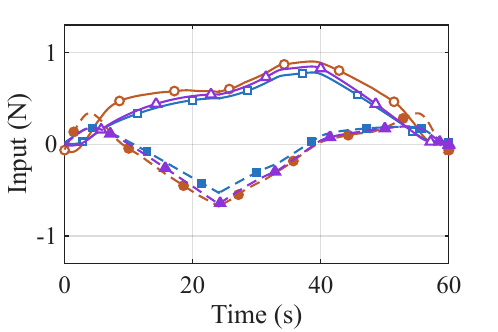}
        \caption{\footnotesize{Rand2}}
    \end{subfigure}
    \begin{subfigure}{0.245\textwidth}
        \includegraphics[width=\linewidth]{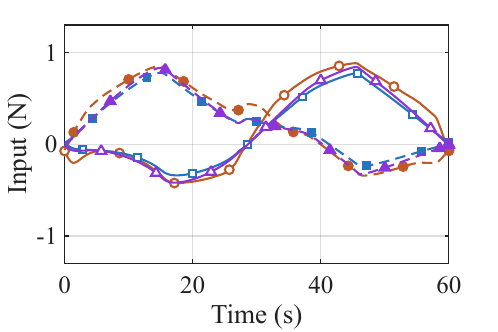}
        \caption{\footnotesize{Rand3}}
    \end{subfigure}
    \begin{subfigure}{0.245\textwidth}
        \includegraphics[width=\linewidth]{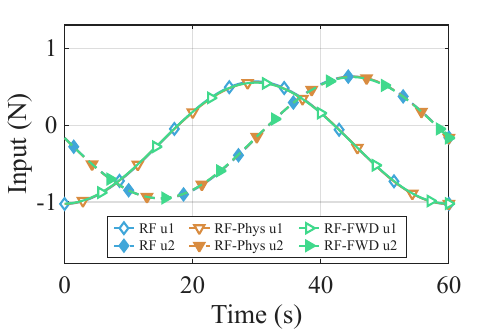}
        \caption{\footnotesize{Circle (RF/RF-Phys/RF-FWD)}}
    \end{subfigure}
    \begin{subfigure}{0.245\textwidth}
        \includegraphics[width=\linewidth]{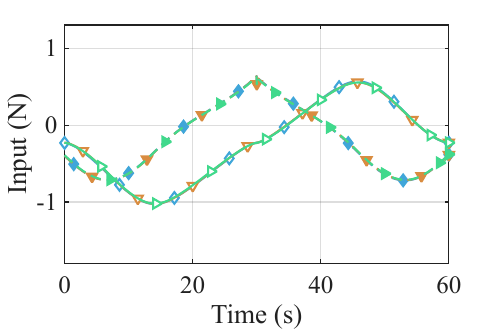}
        \caption{\footnotesize{Heart}}
    \end{subfigure}
    \begin{subfigure}{0.245\textwidth}
        \includegraphics[width=\linewidth]{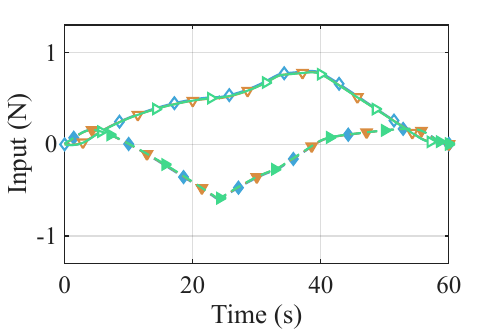}
        \caption{\footnotesize{Rand2}}
    \end{subfigure}
    \begin{subfigure}{0.245\textwidth}
        \includegraphics[width=\linewidth]{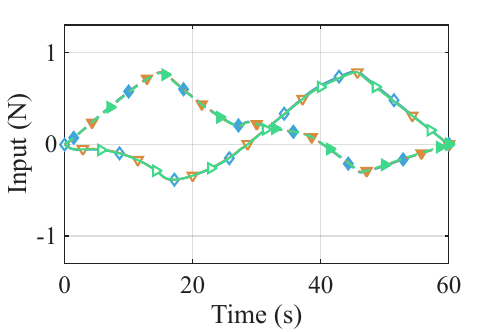}
        \caption{\footnotesize{Rand3}}
    \end{subfigure}
    \caption{\small{Input comparison ($u_1/u_2$). Top: Inverse MLP/LSTM/Transformer. Bottom: RF/RF-Phys/RF-FWD.}}
    \label{fig:2d_inputs}
\end{figure*}

\section{EVALUATION IN SIMULATION}
\label{sec:sim}

In this section, we evaluate the proposed framework in simulation as an open-loop 
inverse dynamics learner, examining whether flow-based modeling 
produces more accurate and smoother actuation than regression-based 
alternatives under dynamic soft-body behavior.
\subsection{System Setup and Data Acquisition}
\label{subsec:sim_setup}

The simulation environment is built on a high-fidelity geometric variable-strain formulation \cite{mathew2025reduced}. 
The robot is a cantilevered cable-driven soft continuum beam with manipulator length $L=0.4$\,m, cross-sectional diameter $D=0.004$\,m, cable distribute distance $W=0.055$\,m, Young's modulus $E=2\times10^9$\,Pa, damping coefficient $\eta=5\times10^7$, and density $\rho=12000$\,kg/m$^3$. Four cables are arranged as two orthogonal antagonistic pairs, and the actuation input is parameterized by differential cable tension,
\begin{equation}
    u_t = [u_1, u_2]^T = [T_{1a}-T_{1b},\, T_{2a}-T_{2b}]^T,
\end{equation}
where $T_{ia}$ and $T_{ib}$ denote the tensions in the $i$-th antagonistic pair. The system state is defined in Cartesian task space as
\begin{equation}
    x_t = [p_x, p_y, p_z, v_x, v_y, v_z]^T,
\end{equation}
where $[p_x,p_y,p_z]^T$ and $[v_x,v_y,v_z]^T$ denote the end-effector position and linear velocity, respectively. The sampling interval is fixed at $\Delta t=0.01$\,s, so the inverse model predicts the actuation associated with each local transition $(x_t,x_{t+1})$.

The training data consist of 3,000 simulated excitation episodes. In each episode, a randomly generated polynomial actuation profile is applied for 5\,s, and the resulting robot motion is recorded to capture its dynamic evolution. Consecutive samples from these trajectories are then converted into local training tuples $(x_t,x_{t+1},u_t)$. All structured evaluation trajectories are excluded from the training data so that the test reflects inverse-dynamics generalization rather than trajectory memorization. All models are trained and tested on the same Apple M4 Max platform.

\subsection{Baseline Comparison}

The test set contains two groups of trajectories: structured geometric trajectories, including circles and character-like paths, and random trajectories. The proposed RF-based inverse-dynamics models are compared with three deterministic baselines: inverse MLP, inverse LSTM, and inverse Transformer. All models predict the same actuation variable $u_t$ \eqref{diff_def}. The inverse MLP takes only the local pair $(x_t,x_{t+1})$ as input, whereas the inverse LSTM and inverse Transformer additionally exploit a longer temporal context. Within the RF family, RF learns the inverse map by conditional generative transport, RF-Physical adds a quasi-static physical prior, and RF-FWD further introduces a forward-dynamics consistency constraint.

\subsubsection{Tracking Accuracy and Computational Cost}

Although flow integration introduces additional overhead, RF inference remains close to 1 ms per step, well within real-time control requirements for soft continuum robots. The base RF and RF-Physical models train within 10 min, substantially faster than the Transformer and LSTM baselines. As shown in Table~\ref{tab:tracking_rmse}, the base RF model already improves substantially over deterministic regression, confirming that generative transport alone benefits inverse-dynamics learning. RF-Physical further reduces random-trajectory RMSE from 5.2 mm to 4.0 mm but offers little gain on structured trajectories, suggesting the quasi-static prior captures dominant elastic behavior yet does not fully account for dynamic transients. RF-FWD achieves the best overall performance (structured: 5.0 mm; random: 3.5 mm), demonstrating that forward-dynamics consistency more effectively constrains inverse prediction than a static prior alone. As shown in Fig.~\ref{fig:tracking_examples}, these improvements also hold qualitatively against deterministic baselines.

\begin{table}[t]
\centering
\caption{\textbf{Open-loop trajectory-tracking RMSE (mm).}}
\label{tab:tracking_rmse}
\renewcommand{\arraystretch}{1.2}
\setlength{\tabcolsep}{0pt}
\begin{tabular*}{\columnwidth}{@{\extracolsep{\fill}} l c c @{}}
\toprule
\textbf{Model} & \textbf{Shape Traj. RMSE} & \textbf{Random Traj. RMSE} \\
\midrule
MLP          & 23.1 & 30.6 \\
LSTM         & 15.6 & 8.4 \\
Transformer  & 13.4 & 12.7 \\
\midrule
RF           & 6.7  & 5.2 \\
RF-FWD       & \textbf{5.0} & \textbf{3.5} \\
RF-Physical  & 6.8  & 4.0 \\
\bottomrule
\end{tabular*}
\end{table}

\subsubsection{Input Quality}

Open-loop inverse dynamics requires not only low tracking error, but also smooth and dynamically consistent actuation. As shown in Fig.~\ref{fig:2d_inputs}, the inverse MLP produces severe high-frequency oscillations, while the inverse LSTM suppresses part of this oscillation at the cost of clear response delay. The inverse Transformer provides a better compromise, but its signals remain less consistent than those generated by the RF-based models. Table~\ref{tab:lag_energy} quantifies control effort using the input energy $E_u=\sum_t \|u(t)\|^2\Delta t$, which serves as an indicator of actuation aggressiveness. Among these models, RF-FWD yields the best overall input quality, achieving the minimum phase lag (45.0\,ms) with low input energy (17.04).

As RF-FWD achieves the superior tracking performance across all test scenarios. Consequently, we select RF-FWD as the representative model for the subsequent stress tests and physical experimental evaluations. 
\begin{table}[t]
\centering
\caption{\textbf{Phase lag and input energy of the generated control signals.}}
\label{tab:lag_energy}
\renewcommand{\arraystretch}{1.2}
\setlength{\tabcolsep}{0pt}
\begin{tabular*}{\columnwidth}{@{\extracolsep{\fill}} l c c @{}}
\toprule
\textbf{Model} & \textbf{Phase Lag (Shape)} & \textbf{Input Energy (Rand)} \\
\midrule
MLP        & 0.0575\,s & 26.47 \\
LSTM         & 0.0775\,s & 16.56 \\
Transformer  & 0.0475\,s & 20.70 \\
RF-FWD       & 0.0450\,s & 17.04 \\
\bottomrule
\end{tabular*}
\end{table}

\subsection{Robustness and Generalization Stress Test}

We further test whether the learned inverse dynamics remains valid beyond the nominal training regime.

\subsubsection{High-Speed Trajectory Tracking}

\begin{figure}
    \centering
    \includegraphics[width=1\linewidth]{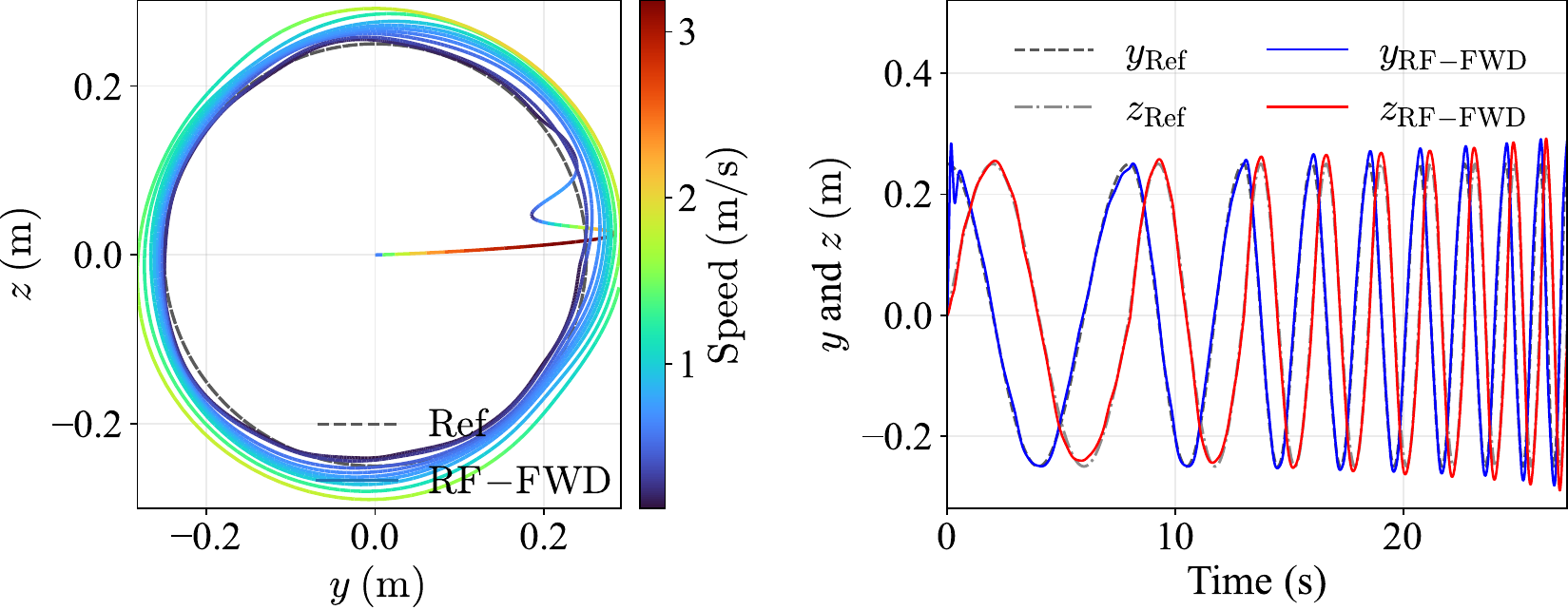}
    \caption{The circular reference trajectories with high speed in simulation.}
    \label{fig:high_speed_simulate}
\end{figure}

A burst circular trajectory is used to stress the model under progressively stronger inertial effects. Inspired by \cite{haggerty2023inertial}, the period of motion is decreased from 5.0\,s to 1.2\,s. As shown in Fig.~\ref{fig:high_speed_simulate}, RF-FWD maintains stable tracking throughout this test and reaches a peak end-effector speed of 2.03\,m/s in simulation. The corresponding $yz$-plane tracking RMSE is 20.65\,mm (5.16\% of the manipulator length). This result shows that the learned inverse map remains effective under substantially faster motion than the nominal training condition.

\subsubsection{Complex Trajectory Tracking}

\begin{figure}
\centering
\begin{minipage}[c]{0.48\linewidth}
    \centering
    \includegraphics[width=\linewidth]{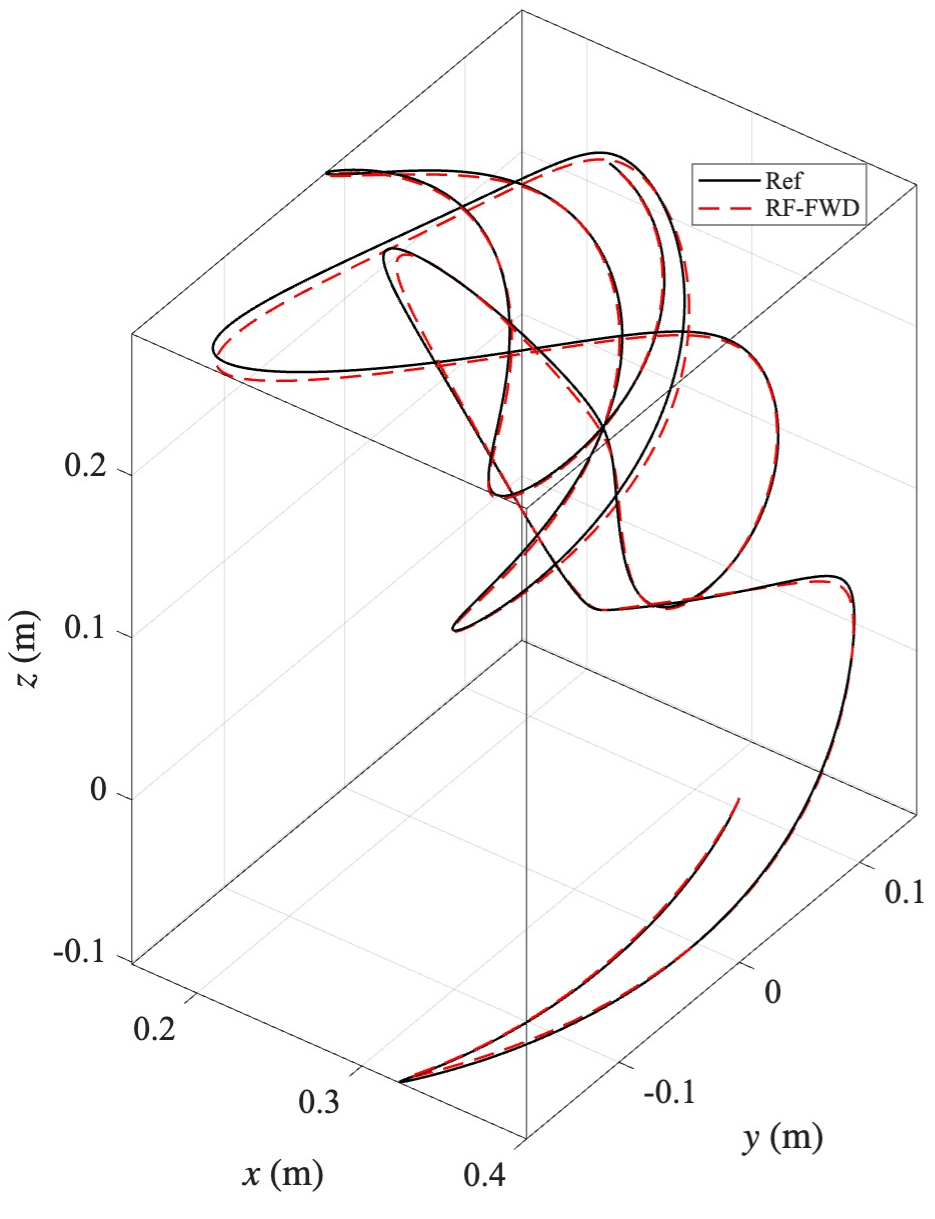}
    \subcaption{3D tip trajectory comparison (Reference vs RF-FWD).}
    \label{fig:fig5_a}
\end{minipage}
\hfill
\begin{minipage}[c]{0.48\linewidth}
    \begin{subfigure}[t]{\linewidth}
        \centering
        \includegraphics[width=\linewidth,trim=2pt 2pt 2pt 1pt,clip]{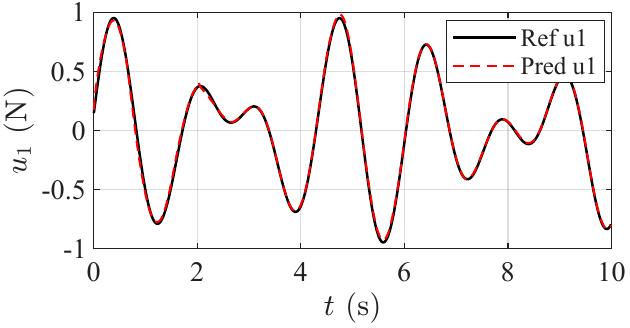}
        \subcaption{Control input $u_1$: reference vs RF-FWD prediction.}
        \label{fig:fig5_b}
    \end{subfigure}
    \vfill
    \begin{subfigure}[t]{\linewidth}
        \centering
        \includegraphics[width=\linewidth,trim=2pt 2pt 2pt 1pt,clip]{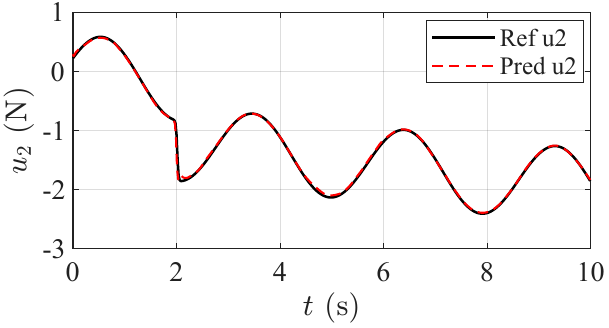}
        \subcaption{Control input $u_2$: reference vs RF-FWD prediction.}
        \label{fig:fig5_c}
    \end{subfigure}
\end{minipage}
\caption{Complex trajectory comparison and input results.}
\label{fig:3d_plot}
\end{figure}

To examine out-of-distribution generalization, the test trajectories are synthesized from a mixture of sinusoidal, sigmoid, and linear functions in the full 3D task space, which differs from the polynomial excitation used during training. A forward-then-inverse protocol is adopted: known actuation inputs $u_t$ are first applied to the simulator to generate a ground-truth trajectory, and the resulting trajectory is then fed to RF-FWD to reconstruct the input sequence. As shown in Fig.~\ref{fig:3d_plot}, the reconstructed inputs closely match the ground truth, achieving a mean absolute error of 0.027 for $u_1$ and 0.033 for $u_2$, corresponding to only 0.34\% and 0.41\% of the full actuation range, respectively. This result indicates that the model captures the underlying inverse dynamics rather than memorizing the polynomial training patterns.

\begin{figure}
	 
    \centering
\includegraphics[width=1\linewidth]{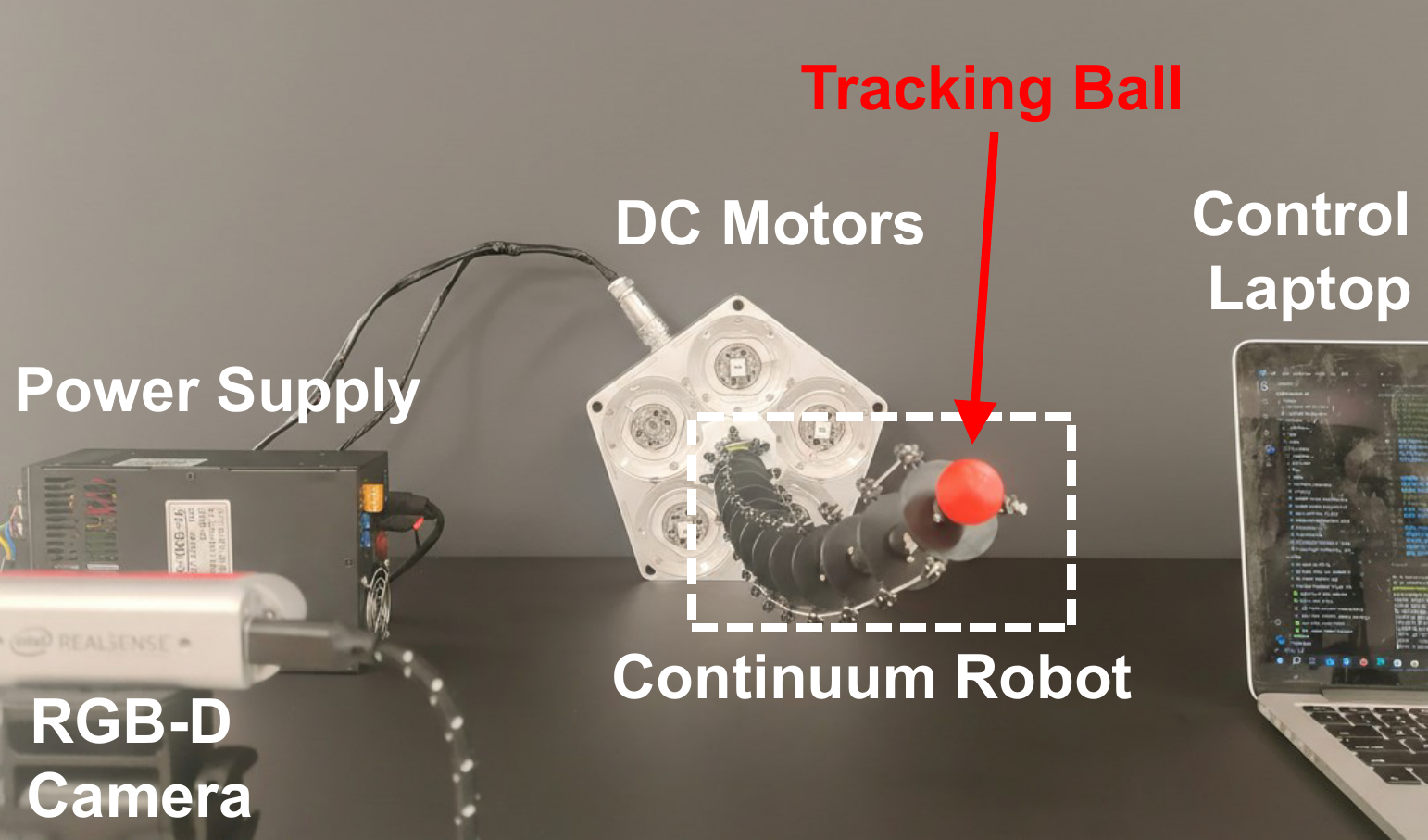}
    \caption{Experimental Setup
}
    \label{fig:exp_setup}
\end{figure}

\begin{figure}
    \centering
    \includegraphics[width=1\linewidth]{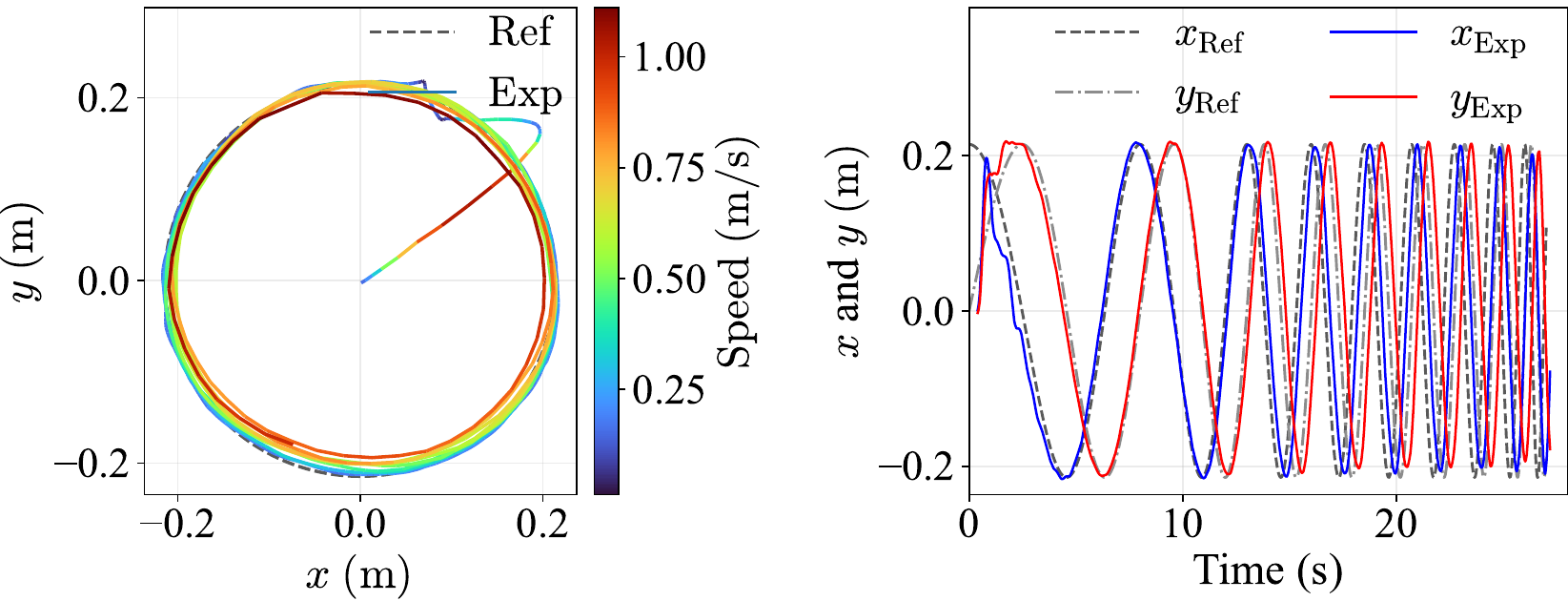}
    \caption{High-speed circular reference tracking in experiments.}
    \label{fig:high_speed_exp}
\end{figure}

\begin{figure*}[!t]
    \centering
    \begin{subfigure}{0.245\textwidth}
        \includegraphics[width=\linewidth]{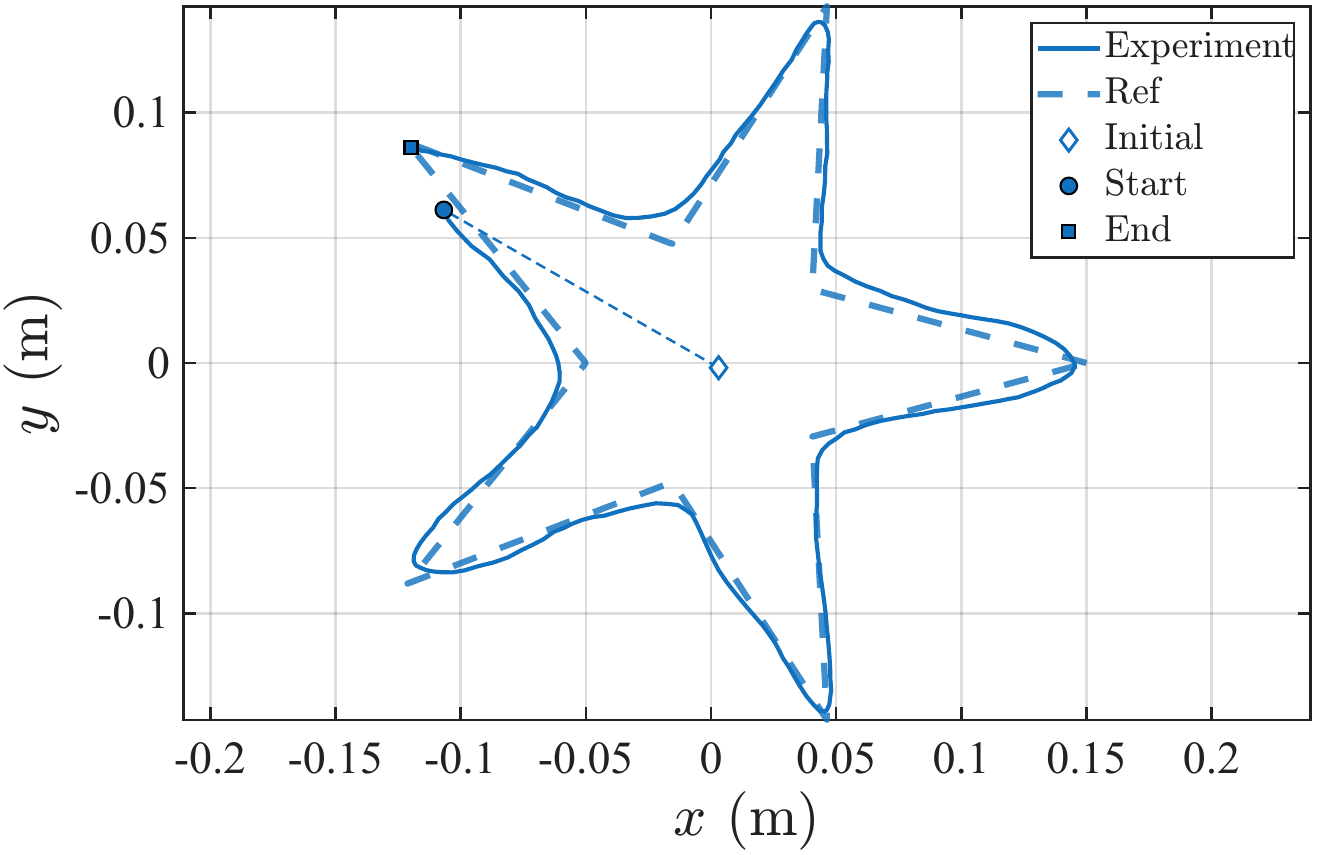}
        \caption{\footnotesize{Star}}
    \end{subfigure}
    \begin{subfigure}{0.245\textwidth}
        \includegraphics[width=\linewidth]{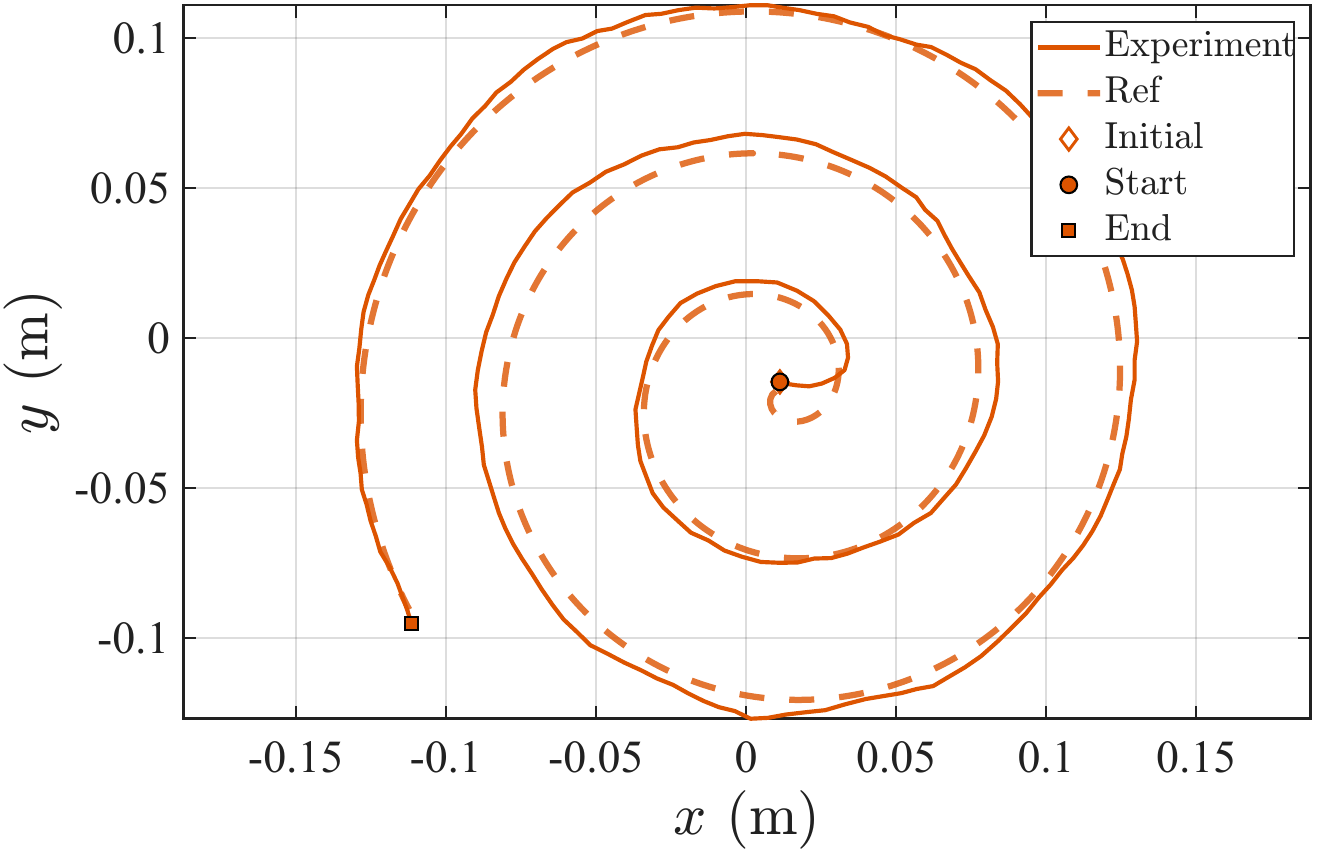}
        \caption{\footnotesize{Spiral}}
    \end{subfigure}
    \begin{subfigure}{0.245\textwidth}
        \includegraphics[width=\linewidth]{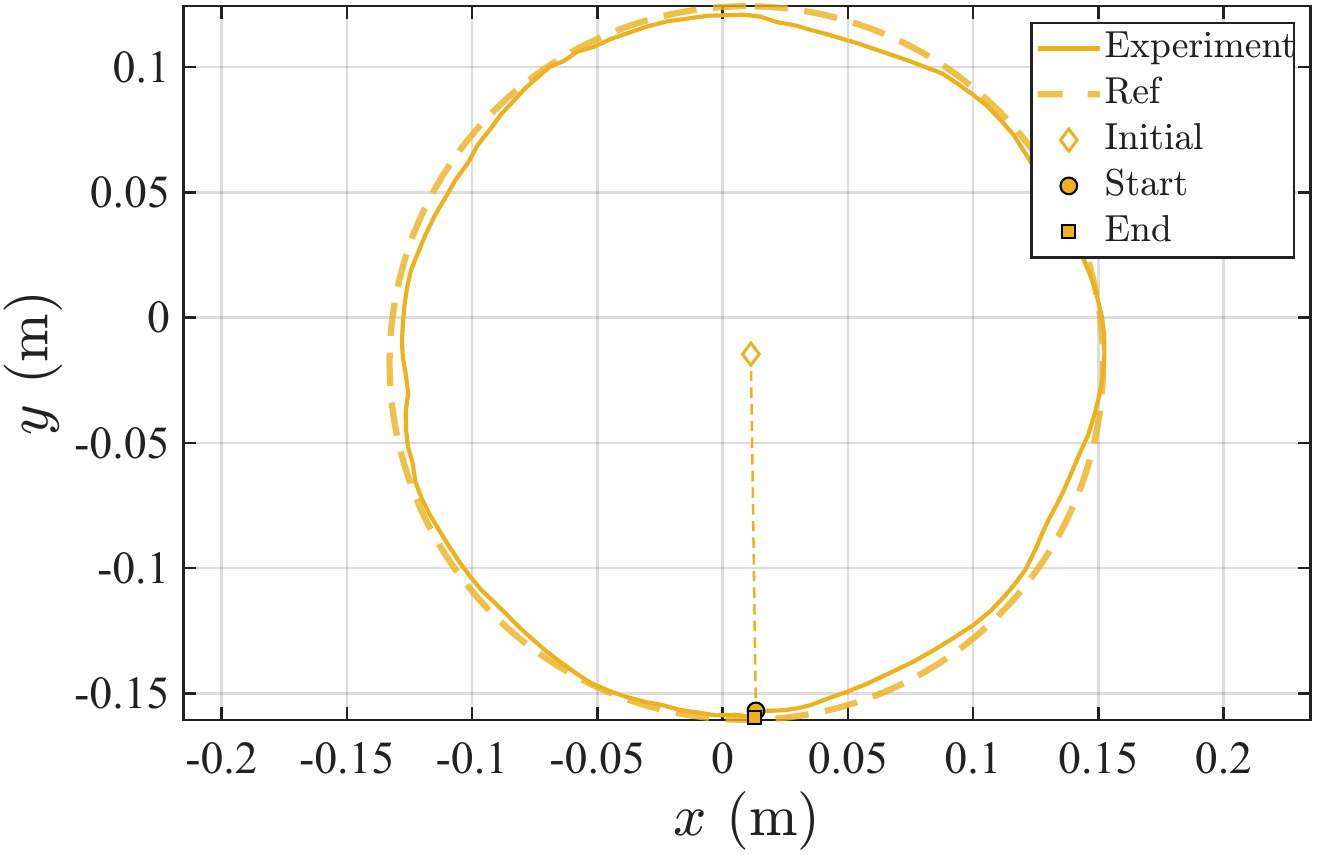}
        \caption{\footnotesize{Circle}}
    \end{subfigure}
    \begin{subfigure}{0.245\textwidth}
        \includegraphics[width=\linewidth]{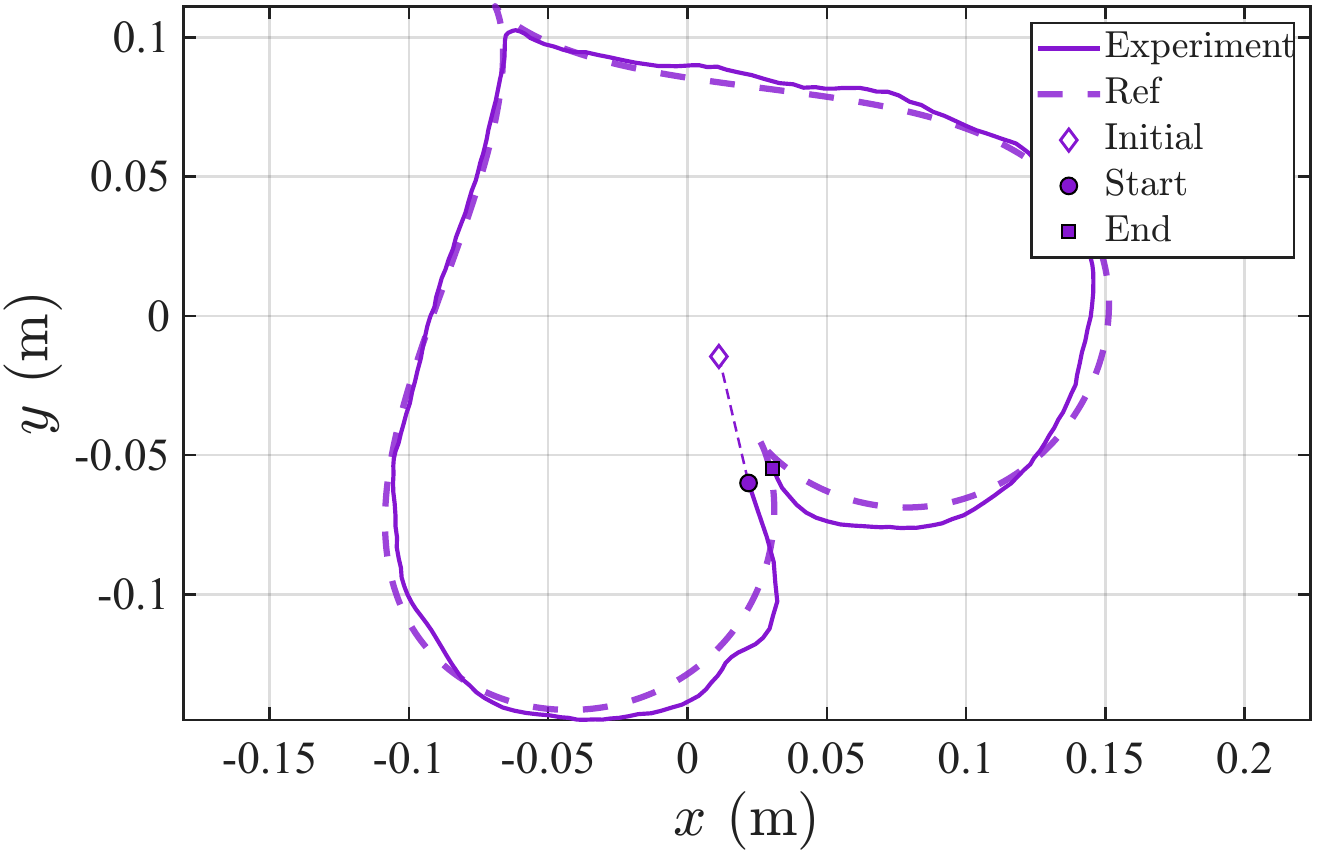}
        \caption{\footnotesize{Heart}}
    \end{subfigure}
    \\
    \begin{subfigure}{0.245\textwidth}
        \includegraphics[width=\linewidth]{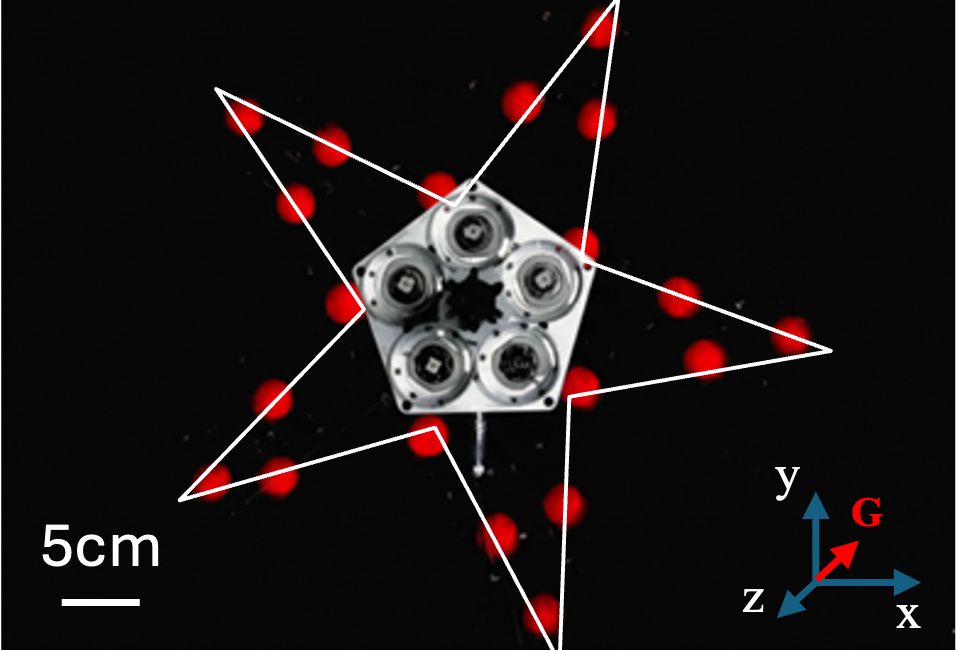}
        \caption{\footnotesize{Star}}
    \end{subfigure}
    \begin{subfigure}{0.245\textwidth}
        \includegraphics[width=\linewidth]{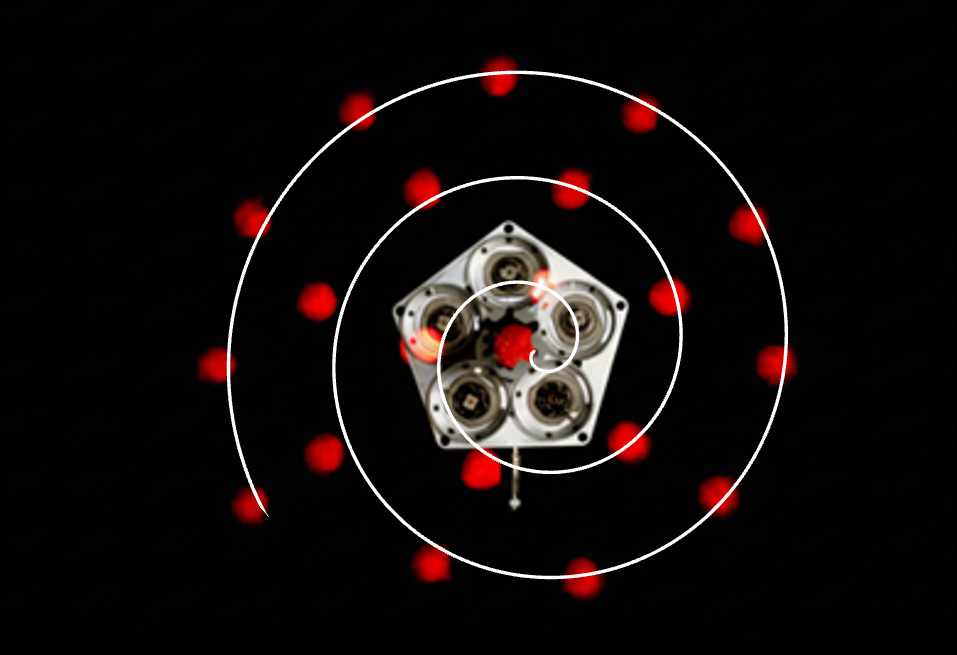}
        \caption{\footnotesize{Spiral}}
    \end{subfigure}
    \begin{subfigure}{0.245\textwidth}
        \includegraphics[width=\linewidth]{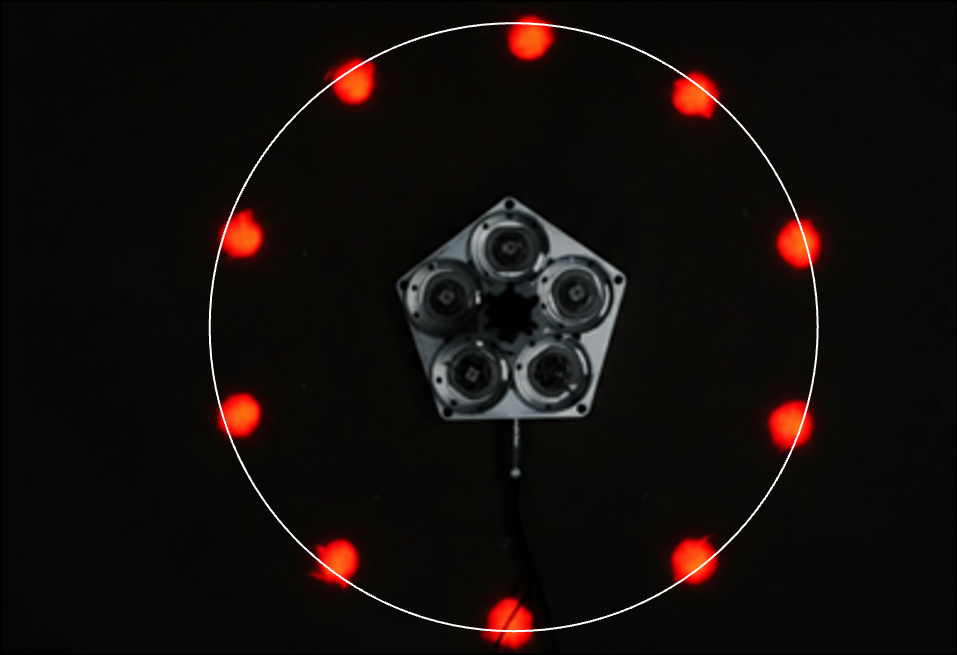}
        \caption{\footnotesize{Circle}}
    \end{subfigure}
    \begin{subfigure}{0.245\textwidth}
        \includegraphics[width=\linewidth]{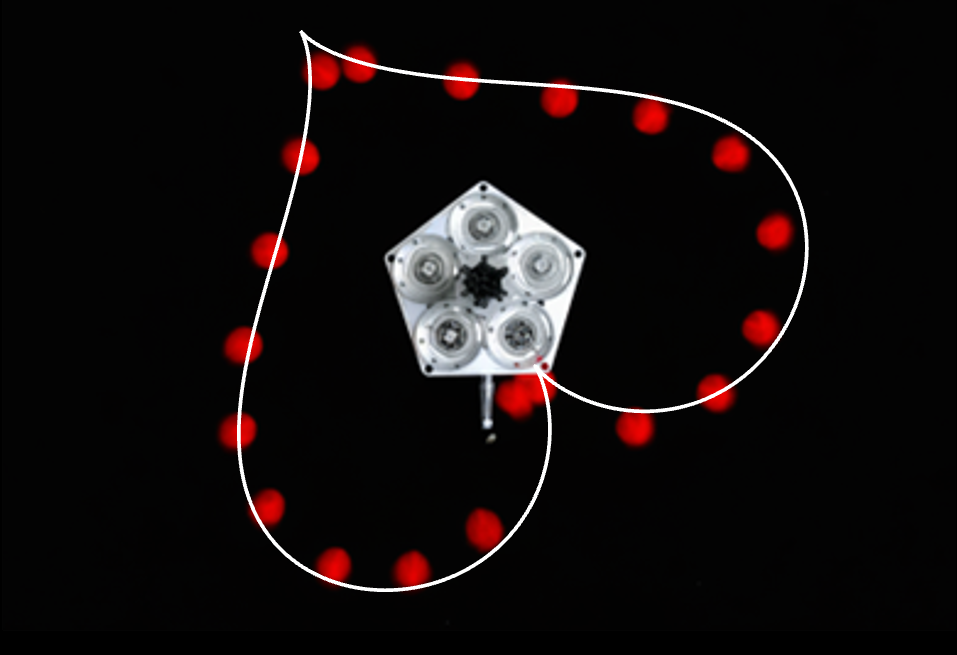}
        \caption{\footnotesize{Heart}}
    \end{subfigure}
    \caption{\small{Vertical plane trajectory tracking performance. (a)-(d) 2D plots of the end-effector position against reference paths. The transient trajectory from the Initial point ($\diamond$) to the Start point ($\bullet$) is excluded from the tracking performance evaluation. (e)-(h) Stroboscopic composite images showing the tracking ball's motion at 0.4 s intervals. Visualization note: Backgrounds in the bottom row were darkened and color-enhanced for contrast.}}
        \label{fig:exp_vertical}
\end{figure*}

\begin{figure*}[!t]
    \centering
    \begin{subfigure}[t]{0.245\textwidth}
        \centering
        \vspace{0pt}
        \includegraphics[width=\linewidth]{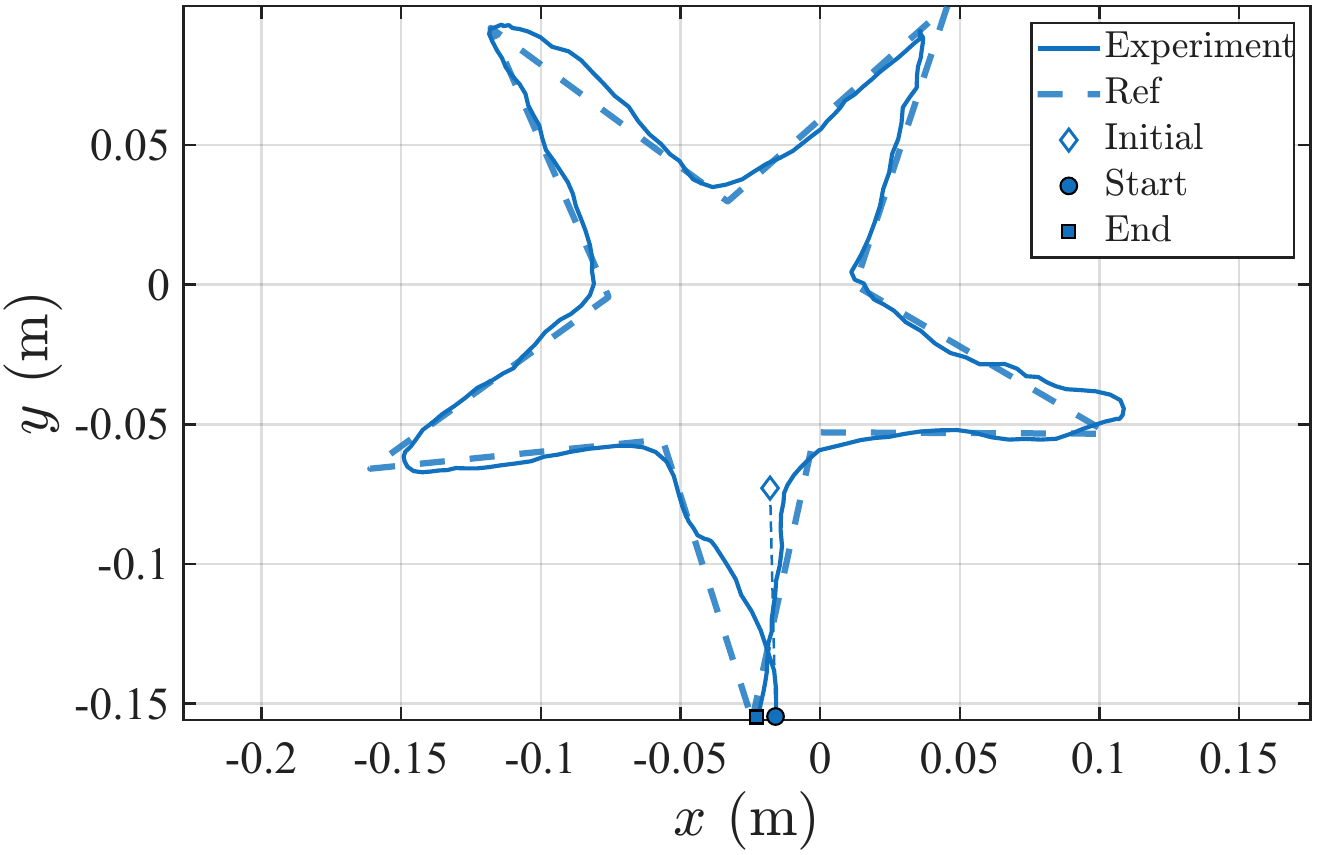}
        \caption{\footnotesize{Star}}
    \end{subfigure}
    \begin{subfigure}[t]{0.245\textwidth}
        \centering
        \vspace{0pt}
        \includegraphics[width=\linewidth]{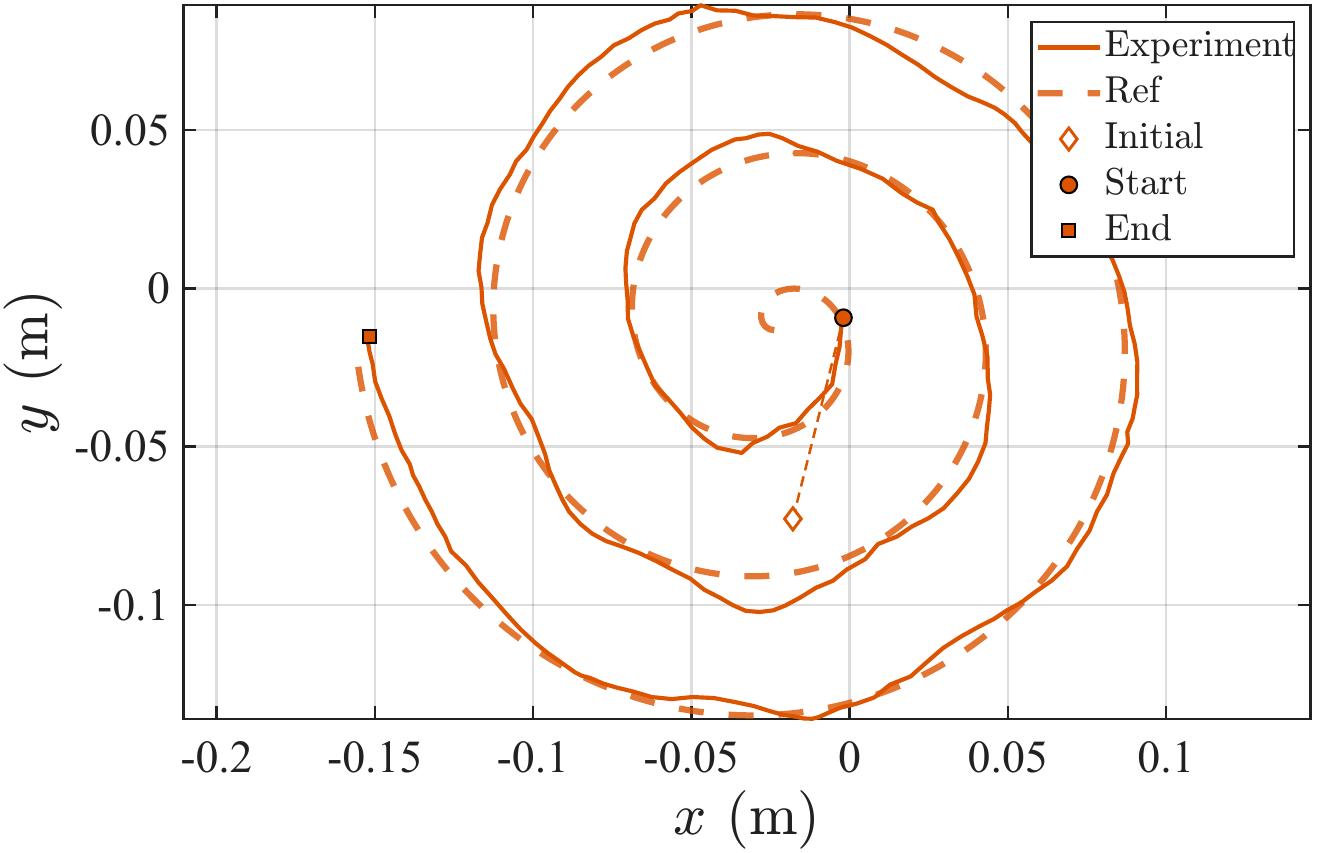}
        \caption{\footnotesize{Spiral}}
    \end{subfigure}
    \begin{subfigure}[t]{0.245\textwidth}
        \centering
        \vspace{0pt}
        \includegraphics[width=\linewidth]{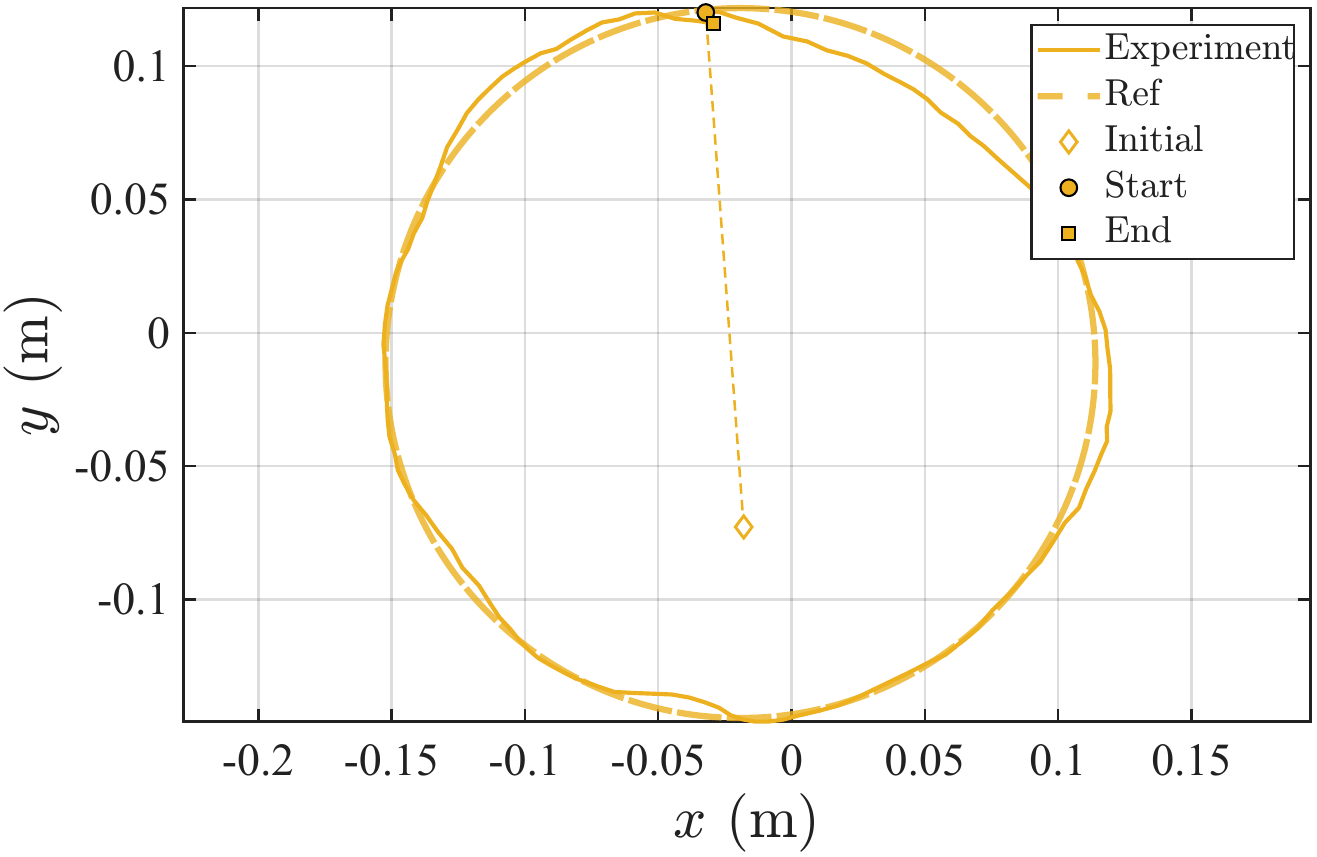}
        \caption{\footnotesize{Circle}}
    \end{subfigure}
    \begin{subfigure}[t]{0.245\textwidth}
        \centering
        \vspace{0pt}
        \includegraphics[width=\linewidth]{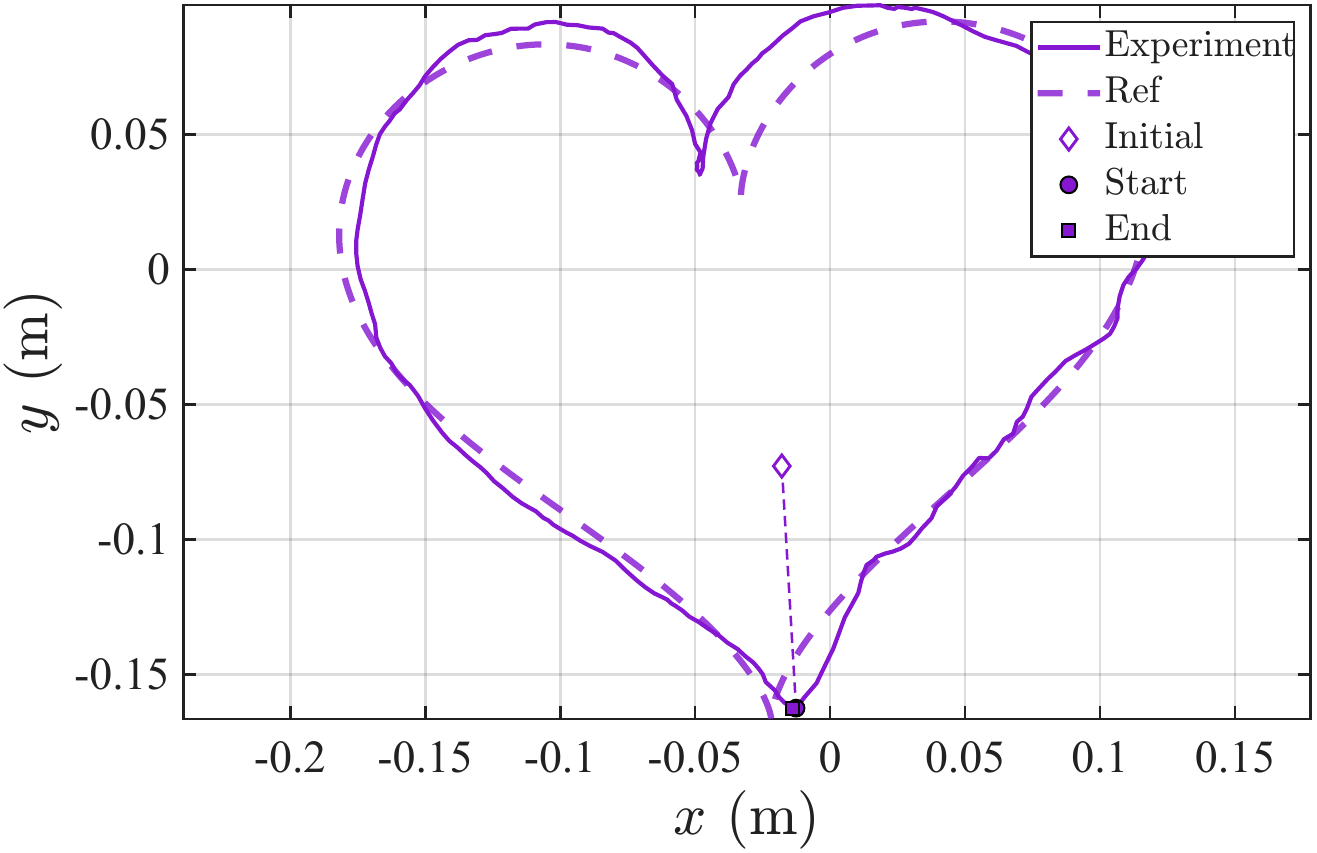}
        \caption{\footnotesize{Heart}}
    \end{subfigure}
    \\
    \begin{subfigure}[t]{0.245\textwidth}
        \centering
        \vspace{0pt}
        \includegraphics[width=\linewidth]{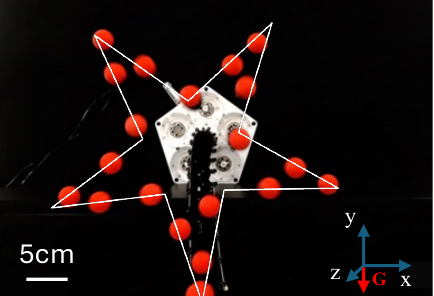}
        \caption{\footnotesize{Star}}
    \end{subfigure}
    \begin{subfigure}[t]{0.245\textwidth}
        \centering
        \vspace{0pt}
        \includegraphics[width=\linewidth]{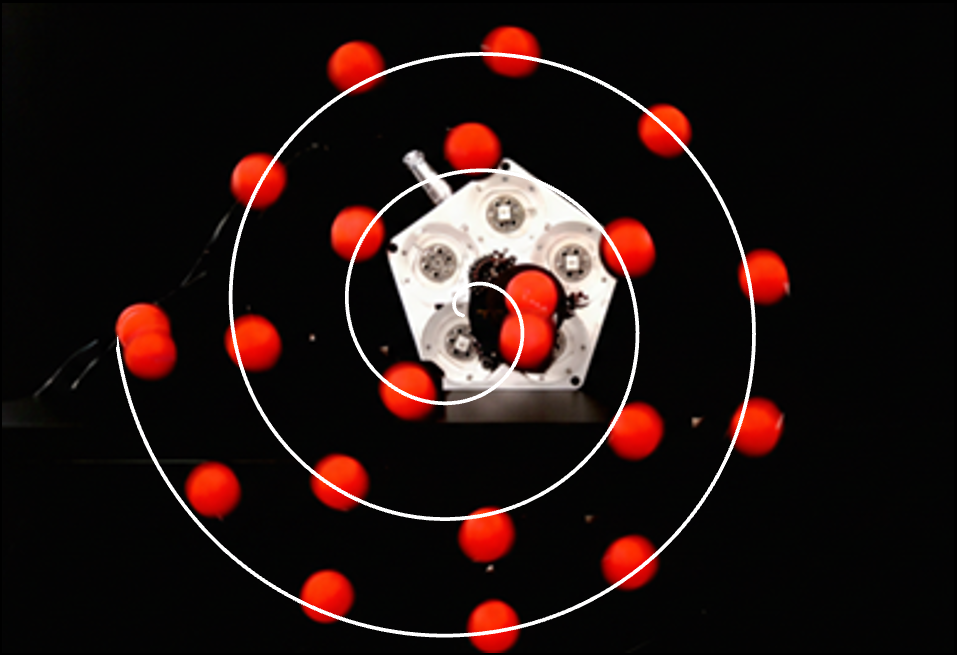}
        \caption{\footnotesize{Spiral}}
    \end{subfigure}
    \begin{subfigure}[t]{0.245\textwidth}
        \centering
        \vspace{0pt}
        \includegraphics[width=\linewidth]{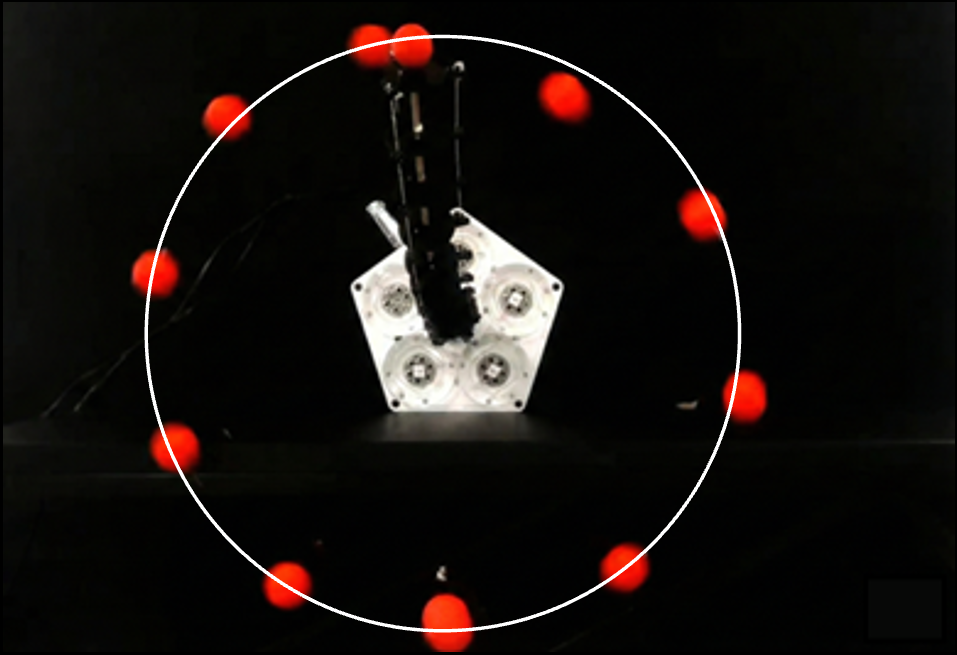}
        \caption{\footnotesize{Circle}}
    \end{subfigure}
    \begin{subfigure}[t]{0.245\textwidth}
        \centering
        \vspace{0pt}
        \includegraphics[width=\linewidth]{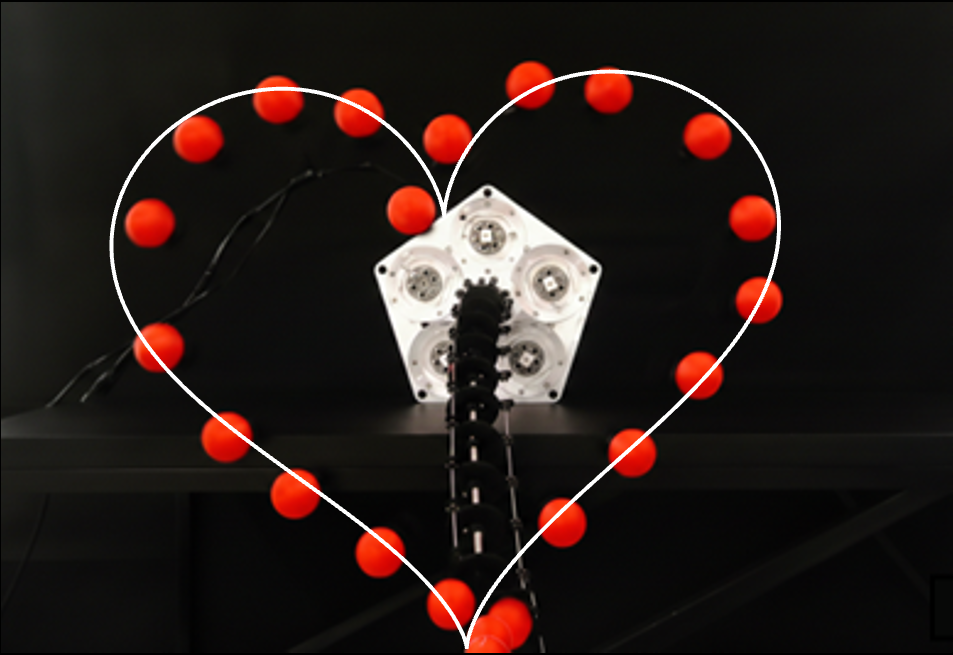}
        \caption{\footnotesize{Heart}}
    \end{subfigure}
    \caption{\small{Horizontal plane trajectory tracking performance. }}
    \label{fig:exp_horizontal}
\end{figure*} 
 
\section{PHYSICAL EXPERIMENTS}
\label{sec:physical}

This section validates the proposed inverse-dynamics framework for open-loop feedforward control on a physical soft continuum robot. Following the description of the hardware platform and deployment pipeline, we present experimental evaluations in two key regimes: high-speed burst tracking and trajectory tracking under varied gravity configurations. To maintain consistency, the core network architecture and training protocols follow the same principles as the simulation phase, with dedicated RF-FWD models adapted for physical configuration under both vertical and horizontal setups.

\subsection{Experimental Setup}
\label{subsec:hardware_setup}

\textit{1) Hardware Platform:}
The experimental platform is a cable-driven soft continuum beam actuated by three cables, as shown in Fig.~\ref{fig:exp_setup}. The robot has length $L=0.45$\,m, cable distribute distance $W=0.025$\,m, diameter $D=0.004$\,m, Young's modulus $E=3.5\times10^9$\,Pa, density $\rho=12400$\,kg/m$^3$, and  damping coefficient $\eta=4\times10^7$ was estimated following \cite{yang2025lightweightdynamicmodelingcabledriven}. A red spherical marker is attached to the end-effector, and its position is measured using an Intel RealSense D435i camera.

\textit{2) Control-Signal Processing for Deployment:}
The RF-FWD model predicts the actuation input $u_t$ at each time step. For motor execution, $u_t$ is converted to cable displacement commands $l_t$ through the simulation model, and the resulting commands are replayed sequentially in open loop. 

\subsection{Experimental Validation}
\label{subsec:results}

\textit{1) High-Speed Burst Tracking:}
To evaluate the dynamic limit on hardware, the burst circular trajectory is tested on the physical robot, with the motion period progressively reduced from 5.0\,s to 1.2\,s, consistent with the simulation protocol. As shown in Fig.~\ref{fig:high_speed_exp}, the robot remains stable throughout the test and reaches a peak end-effector speed of 1.14\,m/s. The corresponding tracking RMSE is 20.65\,mm,  4.59\% of the manipulator length. This result confirms that the learned inverse-dynamics model can be transferred to the physical platform and still maintain stable tracking under high-speed motion.

\textit{2) Trajectory Tracking under Different Gravity Configurations:}
The proposed method is further evaluated in both vertical and horizontal configurations. The horizontal configuration is more challenging because gravity acts asymmetrically on the beam and introduces stronger nonlinear bending and hysteresis. Representative tracking results for the Star, Spiral, Circle, and Heart trajectories are shown in Fig.~\ref{fig:exp_vertical} and Fig.~\ref{fig:exp_horizontal}, and the quantitative errors are summarized in Table~\ref{tab:real_shape_rms}. Millimeter-level tracking accuracy is maintained in both cases, with mean RMSE errors of 4.3\,mm (0.96\% of the robot length) in the vertical configuration and 6.4\,mm (1.42\% of the robot length) in the horizontal configuration. Although the horizontal case produces larger errors, stable open-loop tracking is preserved for all tested patterns.

\subsection{Comparison with Representative Soft-Robot Controllers}
\begin{table}[t]
\centering
\caption{\textbf{Physical tracking RMSE under two gravity configurations (RF-FWD).}}
\label{tab:real_shape_rms}
\renewcommand{\arraystretch}{1.2}
\setlength{\tabcolsep}{0pt}
\begin{tabular*}{\columnwidth}{@{\extracolsep{\fill}} l c c @{}}
\toprule
\textbf{Trajectory} & \textbf{Vertical RMSE (mm)} & \textbf{Horizontal RMSE (mm)} \\
\midrule
Star    & 3.83 & 3.96 \\
Spiral  & 4.94 & 6.52 \\
Circle  & 3.95 & 8.65 \\
Heart   & 4.46 & 6.47 \\
\midrule
\textbf{Mean} & \textbf{4.30} & \textbf{6.40} \\
\bottomrule
\end{tabular*}
\end{table}

 
\begin{table*}[t]
\centering
\footnotesize
\caption{Comparison with representative soft-robot control frameworks.}
\label{tab:comparison_exp}
\renewcommand{\arraystretch}{1.15}
\setlength{\tabcolsep}{3pt}
\begin{tabular*}{\textwidth}{@{\extracolsep{\fill}} l c c c c c @{}}
\toprule
\textbf{Method} & \makecell[c]{\textbf{Ctrl.}\\\textbf{Type}} & \makecell[c]{\textbf{Infer.}\\\textbf{[ms]}} & \makecell[c]{\textbf{Speed [m/s]}\\\textbf{($v/L$ [s$^{-1}$])}} & \makecell[c]{\textbf{Err.}\\\textbf{[mm (\%L)]}} & \makecell[c]{\textbf{Train}\\\textbf{[min]}} \\
\midrule
OL-LSTM \cite{centurelli2021openloop}   & Feedforward & -- & 0.035--0.045 (0.18--0.23) & $\sim$3.2 (1.6\%) & -- \\
Dyn. Oper. \cite{haggerty2023inertial} & Feedback    & -- & 1.50 (3.23) & -- & $<5$ \\
DKoopman \cite{feizi2025deepkoopman}   & Feedback    & 0.44$\pm$0.14 & -- & 1.79$\pm$1.26 (3.25\%) & -- \\
Hyb. MLNN \cite{huang2025hybrid}       & Feedback    & $<3.0$ & 0.005--0.080 (0.03--0.46) & 0.4--1.1 (0.23--0.63\%) & -- \\
PINN-MPC \cite{licher2025adaptive}     & Feedback    & -- & -- & $<3.0$ (2.3\%) & $>11520$ \\
\midrule
{RF-FWD (Ours)} & {Feedforward} & {0.995} & {1.14 (2.53)} & {4.3--6.4 (0.96--1.42\%)} & {20} \\
\bottomrule
\end{tabular*}
\end{table*}

As summarized in Table~\ref{tab:comparison_exp}, RF-FWD is not the best in every individual metric, but it provides a favorable overall trade-off. Compared with the earlier feedforward OL-LSTM controller, RF-FWD improves the normalized dynamic capability from 0.18--0.23\,s$^{-1}$ to 2.53\,s$^{-1}$ while reducing the length-normalized tracking error from 1.6\% to 0.96--1.42\%. Compared with representative feedback controllers, it trades some absolute tracking accuracy for a much simpler feedforward deployment pipeline, while still maintaining millimeter-level experimental accuracy, 0.995\,ms inference time, and only 20\,min training.


\section{CONCLUSION}

This paper presented a flow-matching-based inverse-dynamics model for the open-loop feedforward control of soft continuum robots. By learning the actuation associated with local state transitions and augmenting the inverse model with forward-dynamics consistency and residual physical priors, the proposed framework enabled stable high-frequency open-loop tracking. In simulation and  experiments, RF-FWD reduced trajectory-tracking error by more than 50\% relative to regression baselines, achieved 0.995\,ms inference , and sustained end-effector speeds up to 1.14\,m/s. Future work will extend the framework to multi-segment systems and combine it with lightweight feedback correction and richer state sensing for improved robustness.

\bibliographystyle{IEEEtran}
\bibliography{reference}

\end{document}